\documentclass[preprint,12pt]{elsarticle}

\usepackage[utf8]{inputenc}
\usepackage[T1]{fontenc}
\usepackage{amsmath,amssymb,amsfonts}
\usepackage{graphicx}
\usepackage{grffile} 
\usepackage{booktabs}
\usepackage{multirow}
\usepackage{array}
\usepackage{tabularx}
\usepackage{longtable}
\usepackage{hyperref}
\usepackage{xcolor}
\usepackage{float}
\usepackage{caption}
\usepackage{subcaption}
\usepackage{listings}
\usepackage{textcomp}
\usepackage[margin=1in]{geometry}

\hypersetup{
    colorlinks=true,
    linkcolor=blue,
    citecolor=blue,
    urlcolor=blue
}

\newcolumntype{L}[1]{>{\raggedright\arraybackslash}p{#1}}
\newcolumntype{C}[1]{>{\centering\arraybackslash}p{#1}}
\newcolumntype{R}[1]{>{\raggedleft\arraybackslash}p{#1}}

\begin{document}

\begin{frontmatter}

\title{Exploring parameter-efficient fine-tuning (PEFT) of billion-parameter vision models with QLoRA and DoRA: insights into generalization for limited-data image classification under a 98:1 test-to-train regime}

\author[cornell]{Haiyu Yang}
\author[cornell]{Sumit Sharma}
\author[cornell]{Enhong Liu}
\author[cornell]{Miel Hostens}

\address[cornell]{Cornell University, College of Agriculture and Life Sciences, Ithaca, NY 14853}

\begin{abstract}
Automated behavior classification is essential for precision livestock farming but faces challenges of high computational costs and limited labeled data. This study systematically compared three approaches: training from scratch (ResNet-18, ViT-Small), frozen feature extraction, and parameter-efficient fine-tuning (PEFT) of the DINOv3 foundation model (6.7 billion parameters). We evaluated QLoRA and DoRA across multiple configurations varying rank (8, 16, 64) and target modules (\textit{q\_proj} versus all-linear layers).

With 2,160 verified training images, we assessed generalization of our model on 211,800 test samples, which is essentially a 98:1 test-to-train ratio. Results demonstrated that PEFT substantially outperformed alternatives, where the best QLoRA configuration (all-linear layers and rank=64) achieved 83.16\% test accuracy with only 2.72\% parameters (183.0M) in 5.8 hours, compared to 72.87\% for ResNet-18 (16.8 hours), 61.91\% for ViT-Small (18.7 hours), and 76.56\% for frozen DINOv3 (17.5 hours). DoRA achieved comparable accuracy (83.14\%) but with longer training time (11.0 hours).

Notably, increasing adapter capacity consistently improved generalization while simultaneously not causing overfitting: reducing rank from 16 to 8 decreased test accuracy from 78.38\% to 77.17\%, while expanding from \textit{q\_proj}-only to all-linear layers with rank=64 improved accuracy from 78.38\% to 83.16\%. This suggests underfitting, instead of overfitting, is the primary challenge when adapting foundation models to agricultural imagery. Our findings provide guidelines for deploying billion-parameter vision models with PEFT in agricultural livestock applications.
\end{abstract}


\end{frontmatter}

\section{Introduction}
\label{sec:introduction}

\subsection{Background and Motivation}

Precision livestock farming (PLF) has emerged as a transformative paradigm in modern dairy production, integrating advanced sensing technologies, data analytics, and artificial intelligence to optimize animal health, welfare, and productivity \cite{berckmans2017, jiang2023}. At the core of PLF lies the ability to continuously monitor individual animal behavior, which serves as a powerful indicator of physiological status, health conditions, and welfare states \cite{rohan2024}. As behavioral changes in dairy cattle (e.g., alterations in feeding patterns, locomotion, and social interactions) often precede clinical manifestations of disease, automated behavior recognition is an invaluable tool for early intervention and preventive management \cite{zhang2023}.

The economic significance of automated behavior monitoring cannot be overstated. Lameness alone affects approximately 23--25\% of dairy cows worldwide, resulting in reduced milk yield, delayed reproduction, and increased culling rates \cite{vannuffel2015}. Recent bioeconomic modeling indicates that each lame cow costs approximately \$338 per year on average, with digital dermatitis cases costing about \$431 per cow annually and an additional \$13 per week for each week a cow remains lame \cite{robcis2023}. However, lameness is only one of several economically important conditions that can be identified through behavioral changes. Mastitis is one of the most prevalent diseases in dairy cattle and causes substantial economic losses through reduced milk production, treatment expenses, and discarded milk. The cost of a clinical mastitis case has been estimated to range from about \$200 to more than \$400 depending on severity and stage of lactation \cite{rollin2015}. Transition period metabolic disorders such as ketosis and milk fever also create considerable financial pressure due to reduced production, impaired fertility, and increased veterinary costs \cite{liang2017}. Many of these conditions are associated with measurable behavioral changes, including reductions in feeding time, altered lying behavior, lower locomotion activity, and shifts in daily activity patterns. These changes often occur before clear clinical signs become visible \cite{rutten2013, weigele2018}. Traditional behavior monitoring relies heavily on trained personnel performing visual observations. This approach is labor intensive, subjective, and increasingly difficult to implement as herd sizes continue to expand \cite{warner2020}. As a result, automated monitoring systems that continuously analyze animal behavior offer a scalable approach for early disease detection and improved herd management.

\subsection{Computer Vision for Livestock Behavior Recognition}

The integration of computer vision (CV) and deep learning has revolutionized livestock behavior analysis, enabling continuous, automated monitoring without the need for wearable sensors that may cause animal stress or require maintenance \cite{porto2015}. Early approaches employed convolutional neural networks (CNNs) such as AlexNet, VGG, and ResNet for image-based behavior classification, achieving promising results on constrained datasets \cite{wang2020}. More recently, video-based methods utilizing spatiotemporal architectures, such as 3D CNNs (C3D), ConvLSTM networks, and attention mechanisms, have demonstrated improved performance by capturing temporal dynamics of behavioral patterns. For instance, Qiao et al.\ \cite{qiao2022} combined C3D with ConvLSTM to achieve 90.32\% accuracy in classifying five cattle behaviors (feeding, grooming, walking, standing, and exploring), while Fuentes et al.\ \cite{fuentes2020} integrated frame-level appearance features with spatiotemporal information to recognize 15 hierarchical cattle behaviors with an average accuracy of 85.6\%, demonstrating the effectiveness of temporal feature extraction for complex behavioral recognition tasks.

A critical bottleneck in advancing livestock behavior recognition is the scarcity of available annotated data. Creating large-scale, accurately annotated datasets requires domain expertise from veterinarians or trained ethologists, making data collection time-consuming and expensive \cite{yang2025pipeline}. As a result, most published datasets remain relatively small, limiting their diversity in terms of environments, herd conditions, and animal variability. Models trained under these constraints frequently struggle to generalize to unseen farms or populations, hindering scalability and real-world deployment. Consequently, the central research question is not merely how to design better models, but whether robust and generalizable performance can be achieved under inherently limited annotated data. Therefore, the key challenge is demonstrating that strong and transferable performance can be achieved with limited annotated data. In this context, fine-tuning serves as a practical strategy to overcome data scarcity rather than merely architectural refinement.

\subsection{Foundation Models and Parameter-Efficient Fine-Tuning}

The emergence of foundation models---large-scale models pre-trained on massive datasets that can be adapted to diverse downstream tasks---has fundamentally transformed computer vision research \cite{bommasani2021}. Vision Transformers (ViTs) trained through self-supervised learning paradigms have demonstrated remarkable capability to generate rich, transferable visual representations without requiring labeled data during pre-training \cite{dosovitskiy2021}. Among these, the DINO (Distillation with No Labels) family of models has garnered particular attention for producing features that exhibit emergent semantic understanding and strong performance across diverse visual tasks \cite{caron2021, oquab2024}.

The recently released DINOv3 represents a significant advancement in self-supervised vision models, scaling to 7 billion parameters and training on 1.7 billion unlabeled images \cite{simeoni2025}. DINOv3 introduces several technical innovations, including Gram anchoring to address dense feature degradation during extended training and Rotary Position Embeddings (RoPE) for variable resolution processing. The model achieves state-of-the-art performance across numerous vision benchmarks, often without requiring any task-specific fine-tuning, positioning it as a promising universal vision encoder for specialized domains including agriculture. Despite its capabilities, foundation models like DINOv3 often underperform on highly specialized tasks with domain-specific visual patterns that differ substantially from their pre-training data \cite{wortsman2022, kumar2022}. In agricultural contexts---where images contain unique animal postures, farm environments, and species-specific behaviors---task-specific adaptation through fine-tuning remains necessary to achieve optimal performance \cite{andrew2021, bello2020}.

However, directly fine-tuning such massive models is computationally prohibitive for most research and industry settings. Parameter-Efficient Fine-Tuning (PEFT) methods address this challenge by updating only a small subset of model parameters while keeping the pre-trained weights frozen \cite{houlsby2019}. Low-Rank Adaptation (LoRA) has emerged as the dominant PEFT approach, introducing trainable low-rank matrices that approximate weight updates with dramatically reduced memory and computational requirements \cite{hu2022}. QLoRA extends this framework by quantizing the pre-trained model to 4-bit precision, further reducing memory footprint and enabling fine-tuning of billion-parameter models on consumer-grade GPUs \cite{dettmers2023}.

More recently, Weight-Decomposed Low-Rank Adaptation (DoRA) has been proposed to close the performance gap between PEFT methods and full fine-tuning \cite{liu2024dora}. DoRA decomposes pre-trained weights into magnitude and directional components and updates them independently, enabling a training pattern that closely mimics full fine-tuning while remaining parameter efficient. DoRA has demonstrated consistent improvements over LoRA across language and vision tasks, achieving state-of-the-art PEFT results \cite{liu2024dora}.

Despite the success of foundation models and PEFT methods in natural language processing and general computer vision, their application to agricultural domains remains limited. A recent study by Espejo-Garc\'ia et al.\ \cite{espejo2025} demonstrated the potential of DINOv2 with LoRA fine-tuning for crop disease and weed identification, highlighting the transferability of foundation model features to agricultural imagery. However, systematic evaluations of these approaches for livestock behavior classification are absent from the literature.

\subsection{Research Gap and Objectives}

While considerable progress has been made in livestock behavior recognition using deep learning, several critical gaps remain unaddressed. First, most existing studies train models from scratch on domain-specific datasets, failing to leverage the rich representations learned by modern foundation models. Second, the few studies that employ pre-trained models typically use frozen feature extraction, which may not adequately adapt features to the unique visual characteristics of agricultural imagery. Third, there is a notable absence of systematic comparisons between different behavior detection strategies with CV in the agricultural sector---including training from scratch, frozen features, and PEFT methods---under controlled experimental conditions.

Furthermore, a persistent challenge in agricultural AI research is the discrepancy between model performance on curated validation sets and generalization to large-scale, real-world data. Many reported accuracy figures are based on test sets that share similar characteristics with training data, potentially overestimating practical deployment performance. Rigorous evaluation frameworks that assess generalization across substantially larger and more diverse test sets are essential for translating research advances into operational systems.

This study addresses these gaps through the following objectives:

\begin{enumerate}
    \item \textbf{Large-scale generalization assessment:} we evaluate model performance on a test set nearly 100 times larger than the training set (211,800 vs.\ 2,160 samples), providing a rigorous assessment of generalization capability that reflects real-world deployment scenarios;
    \item \textbf{Systematic comparison of training paradigms:} we compare three distinct approaches including training from scratch using ResNet-18 and ViT-Small, frozen feature extraction with DINOv3, and PEFT of DINOv3 for nine-class dairy cow behavior classification;
    \item \textbf{Comprehensive evaluation of PEFT hyperparameters:} we systematically evaluate QLoRA and DoRA across multiple configurations, varying rank values (8, 16, 64) and target module selections (\textit{q\_proj} only vs.\ all-linear layers), to identify optimal settings for agricultural imagery; and
    \item \textbf{Practical deployment guidelines:} based on our findings, we provide actionable recommendations for applying foundation models and PEFT methods to livestock monitoring applications, considering trade-offs between accuracy, computational resources, and training time.
\end{enumerate}

\subsection{Contributions}

The principal contributions of this study are summarized as follows:

\begin{enumerate}
    \item We demonstrate that a small, carefully curated dataset of 2,160 verified images can generalize effectively to classify over 211,000 samples when combined with appropriate PEFT strategies, providing evidence that dataset quality can compensate for quantity in specialized domains;
    \item We provide a systematic comparison of QLoRA and DoRA for adapting billion-parameter vision models to agricultural imagery, revealing that both techniques achieve comparable performance with the optimal configuration (all-linear layers, rank=64) reaching approximately 83\% test accuracy;
    \item We present the first application of DINOv3, the latest self-supervised vision foundation model released in August 2025, to livestock behavior classification, demonstrating its superior feature representations compared to training from scratch;
    \item We identify and explain a counterintuitive finding that increasing adapter capacity (rank and target modules) improves generalization rather than causing overfitting, suggesting that the primary challenge in adapting foundation models to agricultural tasks is underfitting rather than overfitting;
    \item We provide practical guidelines for practitioners seeking to deploy foundation vision models for livestock monitoring, including recommendations for hyperparameter selection, computational resource requirements, and dataset curation strategies.
\end{enumerate}

\section{Materials and Methods}
\label{sec:methods}

\subsection{Dataset Verification and Curation}

Our dataset consisted of nine dairy cow behaviors: Drinking, Eating head down, Eating head up, Lying, Standing, Walking, Frontal pushing, Gallop, and Leap. We initially extracted 300 images per behavior class collected from 2 available datasets namely MMCows (\url{https://github.com/neis-lab/mmcows}) and PlayBehaviour from Yang et al.\ \cite{yang2026play} (\url{https://github.com/Bovi-analytics/Individual-Behavior-Analysis-with-CV}). Rather than using these images directly, we implemented a comprehensive manual verification process combining veterinary expertise and computer vision principles to ensure both label accuracy and representational diversity. It is important to notice that in the original MMCows dataset, eating behavior is addressed as ``feeding'', which refers to a different behavior in the agricultural field that consists of human interaction. Therefore, we will refer to ``feeding'' as ``eating'' in the content of this paper.

\subsubsection{Verification Protocol}

Two annotators with veterinary training background and experience in animal behavior assessment reviewed each candidate image. The verification process evaluated several criteria:

\textbf{Behavioral accuracy:} We confirmed that each image clearly represented the intended behavior class. We focused on overall body configuration, spatial relationships, and contextual indicators rather than fine anatomical details that might not be visible in typical farm surveillance conditions. For example, for walking behavior we assessed asymmetric leg positions or forward movement indicators rather than requiring clear visualization of individual hoof placement. For eating behaviors, we distinguished head down from head up postures by evaluating muzzle position relative to shoulder height and animal orientation toward eating areas. Table~\ref{tab:verification_criteria} presents the specific anatomical and postural criteria used to verify each behavior class during the manual review process.

\begin{table}[!htbp]
\centering
\caption{Criteria developed for manual verification of dairy cow behaviors in this study.}
\label{tab:verification_criteria}
\small
\begin{tabularx}{\textwidth}{L{1.6cm}L{3.2cm}L{2.8cm}L{3.2cm}L{3.2cm}}
\toprule
\textbf{Behavior Class} & \textbf{Primary Anatomical Criteria} & \textbf{Postural Requirements} & \textbf{Spatial Relationships} & \textbf{Additional Verification Points} \\
\midrule
Drinking & Muzzle oriented toward and in proximity to water source & Head lowered with neck extended forward & Animal positioned near water trough or bowl & Water source visible in frame; head position distinctly lower than eating postures \\
\addlinespace
Eating head down & Muzzle positioned below shoulder height & Head angled downward toward feed surface & Body at or near feed bunk or eating area & Head clearly directed toward ground or feed bunk; neck in downward curve \\
\addlinespace
Eating head up & Muzzle at or above shoulder height & Head elevated while at eating location & Body positioned at feed bunk area & Head raised above feed surface; distinct from standing by proximity to feed area \\
\addlinespace
Lying & Torso in contact with ground surface & Body horizontal and resting on ground or bedding & Significant vertical gap absent between body and ground & Legs not in standing position; body clearly supported by ground \\
\addlinespace
Standing & Legs in vertical weight-bearing position & Upright posture with torso elevated & Clear space visible between ventral body surface and ground & Four legs supporting body weight; stationary or minimal movement \\
\addlinespace
Walking & At least one leg raised or in forward motion & Body in upright position with leg displacement & Forward movement trajectory evident & Leg position asymmetric or mid-stride; motion blur may be present \\
\addlinespace
Frontal pushing & Two animals with heads in close contact & Heads pressed together; neck extended & Two animals facing each other at close range & Head-to-head positioning; typically at feed barrier \\
\addlinespace
Gallop & Legs in dynamic extended position & Body showing motion with pronounced leg extension & Rapid movement indicators present & Multiple legs off ground; motion blur common; high energy evident \\
\addlinespace
Leap & Legs elevated or body airborne & Body showing upward or forward trajectory & Animal distinctly above normal standing height & Vertical displacement visible; body appears suspended \\
\bottomrule
\end{tabularx}
\end{table}

Images depicting transitional states were excluded. For example, we removed images showing cows in the process of lying down or standing up, or animals with heads at intermediate positions between eating postures. This decision prioritized classification clarity over dataset size, as ambiguous labels would introduce noise that could impair model training.

\textbf{Visibility and occlusion patterns:} Commercial dairy facilities present numerous occlusions from infrastructure (feed barriers, stall dividers, waterers), other animals, and equipment. We retained images with varying occlusion levels to ensure models could handle real-world conditions. This included partial occlusions (20--40\% of body obscured), and challenging viewing angles where certain body parts required inference from visible segments.

\textbf{Environmental context:} We examined whether images contained environmental cues that might lead to spurious correlations. In the wild, ``Drinking'' naturally occurs near water sources, while ``Eating head down'' appears predominantly at feed bunks. To test whether models learned actual behavioral features versus location shortcuts, we included images where environmental context was less obvious or atypical.

\textbf{Image quality variation:} Unlike typical dataset curation that favors high-quality images, we intentionally retained images with quality degradation to increase robustness. This included low-resolution captures, motion blur from animal movement, variable lighting (overexposure from sunlight, underexposure in dim areas, high-contrast shadows), and compression artifacts from video encoding.

\subsubsection{Variance Enhancement Strategy}

To maximize dataset diversity and test model generalization, we deliberately selected images that increased classification difficulty:

\textbf{Camera perspectives:} We included unconventional viewing angles; extreme oblique angles nearly parallel to the floor, elevated overhead views that compressed spatial relationships, low-angle shots near ground level, and variable distances from close-ups to wide shots.

\textbf{Animal density:} We also included crowded scenarios where behavior classification becomes more challenging: tightly packed feeding areas with overlapping bodies, lying areas with minimal spacing between animals, and group contexts where target animals were surrounded by others performing different behaviors.

\textbf{Temporal variation:} Images spanned different times of day to capture natural lighting changes, variable coat conditions (wet from cleaning systems, seasonal shedding), and facility appearance variations.

\textbf{Individual variation:} All animals in the dataset were Holsteins; however, substantial individual variation was still present. The dataset includes animals at different life stages, with both calves and adult cows represented for certain behaviors. For most behaviors, we collected multiple clear, high-quality examples from the same animal IDs to ensure reliable behavior representation. At the same time, we intentionally included as many different animals as possible rather than repeatedly using the same individuals for the same behaviors.

\textbf{Farm heterogeneity:} Images were collected from different areas within farms, capturing animals in a range of locations and settings. This spatial diversity helped prevent models from learning location- or farm-specific visual cues rather than focusing on their behaviors themselves.

\subsubsection{Final Dataset Composition}

The original dataset consisted of 300 images per behavior class, totaling $9 \times 300 = 2{,}700$ images. After verification, we retained 240 images per class (80\%) for training and 60 images per class (20\%) for validation, resulting in a total of 2,160 training images and 540 validation images. This relatively small training set was intentional. Our goal was to evaluate whether foundation models using parameter-efficient fine-tuning could generalize effectively from a limited but carefully curated dataset.

During dataset verification, we prioritized label accuracy and meaningful diversity over increasing dataset size, resulting in a challenging benchmark to test whether high-quality data can compensate for limited quantity when adapting large pre-trained models. Final model performance was assessed on a substantially larger test set of 211,800 samples from the original uncurated dataset, including 209,881 samples from the MMCows source and 1,919 samples from the PlayBehavior source.

\subsection{Data Augmentation}

To address the limited size of the training dataset and improve model generalization, we implemented a comprehensive data augmentation pipeline using the Albumentations library \cite{buslaev2020}. Each training image was augmented with a $3\times$ multiplier, expanding the effective training set from 2,160 to 6,480 samples. Augmentations were applied exclusively to the training data using medium-intensity transformations, whereas validation and test images were subjected only to resizing and normalization to preserve evaluation consistency.

The augmentation pipeline consisted of four categories of transformations. Geometric transformations included horizontal flipping ($p=0.5$), random rotation within $\pm 15$ degrees ($p=0.5$), shift-scale-rotate operations with 10\% shift limit and 80--120\% scale range ($p=0.5$), and perspective distortion with 5--10\% scale ($p=0.3$).

Color-based augmentations were applied using a one-of selection ($p=0.8$) from random brightness and contrast adjustment ($\pm 20\%$), hue-saturation-value shifts (hue $\pm 10$, saturation $\pm 20$, value $\pm 20$), and color jittering with brightness and contrast variation of $\pm 20\%$, saturation $\pm 20\%$, and hue $\pm 10\%$.

To account for sensor noise and motion-related artifacts, noise and blur transformations were applied using a \textit{one-of} selection ($p=0.3$) from Gaussian noise (variance 10--50), Gaussian blur (kernel 3--5), and motion blur (kernel size 5). Additionally, CoarseDropout was applied ($p=0.2$) with 1--8 rectangular holes of size up to 10\% of the image dimension, simulating occlusions commonly encountered in barn environments.

All images were resized to $224 \times 224$ pixels and normalized using ImageNet statistics (mean$=[0.485, 0.456, 0.406]$, std$=[0.229, 0.224, 0.225]$) to ensure compatibility with pre-trained vision models.

\subsection{Experimental Approaches}

We compared three distinct approaches for cow behavior classification: training from scratch, frozen feature extraction, and parameter-efficient fine-tuning.

\subsubsection{Baseline: Training from Scratch}

Two neural network architectures were trained from scratch with randomly initialized weights to establish baseline performance. The first architecture, ResNet-18 \cite{he2016}, is a convolutional neural network with 18 layers utilizing residual connections. The model contains 11.2 million parameters with a 512-dimensional feature space before the classification head. For this study, we modified the final fully connected layer to output 9 classes corresponding to our defined behavior categories.

The second architecture, ViT-Small \cite{dosovitskiy2021}, is a Vision Transformer with small configuration (patch size 16, embedding dimension 384). The model contains 21.7 million parameters and was implemented using the timm library \cite{wightman2019}. Like ResNet-18, the original classification head was replaced with a linear layer outputting 9 classes.

Both models were trained for 150 epochs using the AdamW optimizer \cite{loshchilov2019} with weight decay of 0.01. Due to differences in architectural characteristics, ResNet-18 was trained with a learning rate of $1 \times 10^{-3}$, whereas ViT-Small employed a smaller learning rate of $5 \times 10^{-5}$ to account for the higher sensitivity of transformer-based models to learning rate selection. A cosine annealing learning rate schedule was applied, and label smoothing of 0.1 was used for regularization. Training was conducted with batch size 32 on a Tesla V100-PCIE-16GB GPU.

\subsubsection{Frozen Feature Extraction}

For the frozen feature extraction approach, we employed DINOv3 \cite{simeoni2025}, a self-supervised vision foundation model based on the ViT-Giant architecture with 6.7 billion parameters. The specific model variant (\texttt{facebook/dinov3-vit7b16-pretrain-lvd1689m}) was pre-trained on 1.7 billion images using distillation with no labels and produced 4096-dimensional embeddings.

To reduce memory consumption, the DINOv3 backbone was loaded in half-precision (float16) and kept completely frozen during training. A lightweight classification head was attached, consisting of three fully connected layers: Linear($4096 \to 1024$) with ReLU activation and 30\% dropout, Linear($1024 \to 512$) with ReLU activation and 15\% dropout, and Linear($512 \to 9$) for final classification. Only the classification head (4.7 million parameters, 0.07\% of the full model) was trained.

Training was conducted for 80 epochs using AdamW optimizer with learning rate $1 \times 10^{-3}$, weight decay 0.01, and cosine annealing schedule. Due to memory constraints from the large backbone, batch size was reduced to 8 with gradient accumulation over 4 steps, yielding an effective batch size of 32.

\subsubsection{Parameter-Efficient Fine-Tuning}

To adapt DINOv3 to the cow behavior classification task in a parameter-efficient manner, we evaluated two parameter-efficient fine-tuning (PEFT) strategies: Quantized Low-Rank Adaptation (QLoRA) and Weight-Decomposed Low-Rank Adaptation (DoRA).

QLoRA \cite{dettmers2023} combines 4-bit quantization of pre-trained weights with low-rank adaptation. The DINOv3 backbone was quantized using 4-bit NormalFloat (NF4) quantization with double quantization enabled to further reduce memory footprint. Low-rank adapter matrices were then injected into the frozen quantized backbone, enabling gradient flow through the adapters while the base weights remained fixed. The compute dtype was set to float16 for efficient forward and backward passes.

DoRA \cite{liu2024dora} extends LoRA by decomposing pre-trained weights into magnitude and directional components, updating them independently during fine-tuning. This decomposition allows DoRA to more closely approximate full fine-tuning behavior while maintaining parameter efficiency. DoRA was implemented by setting the \texttt{use\_dora} flag in the PEFT library's \texttt{LoraConfig}.

For both techniques, we systematically evaluated multiple configurations varying in rank and target modules. We tested rank values of $r \in \{8, 16, 64\}$ with corresponding alpha values of $\alpha = 2r$ (i.e., $\alpha \in \{16, 32, 128\}$). For target modules, we compared two adapter configurations: the \textit{q\_proj} only setting and the \textit{all-linear} setting as shown in Table~\ref{tab:peft_configs}. The \textit{q\_proj} only setting restricts adapters to the query projection layers in the self-attention blocks, resulting in minimal parameter overhead. The \textit{all-linear} setting extends adapters to all linear transformations in the transformer, including attention projections (Q, K, V, and output) and MLP layers, enabling broader adaptation at the cost of increased parameter count.

The same classification head architecture used in the frozen feature extraction was attached to the PEFT-adapted backbone. LoRA dropout was set to 0.05, and adapters were configured with bias=``none'' to minimize additional parameters.

\begin{table}[!htbp]
\centering
\caption{QLoRA and DoRA configurations evaluated in this study.}
\label{tab:peft_configs}
\small
\begin{tabular}{llcccc}
\toprule
\textbf{Method} & \textbf{Target Modules} & \textbf{Rank ($r$)} & \textbf{Alpha ($\alpha$)} & \textbf{Trainable Params} & \textbf{\% of Total} \\
\midrule
QLoRA & q\_proj     & 8  & 16  & 2.6M   & 0.04\% \\
QLoRA & q\_proj     & 16 & 32  & 5.2M   & 0.08\% \\
QLoRA & all-linear  & 16 & 32  & 46.8M  & 0.70\% \\
QLoRA & all-linear  & 64 & 128 & 183.0M & 2.72\% \\
\midrule
DoRA  & q\_proj     & 8  & 16  & 2.8M   & 0.04\% \\
DoRA  & q\_proj     & 16 & 32  & 5.4M   & 0.08\% \\
DoRA  & all-linear  & 16 & 32  & 48.4M  & 0.72\% \\
DoRA  & all-linear  & 64 & 128 & 184.5M & 2.75\% \\
\bottomrule
\end{tabular}
\end{table}

\subsection{Implementation Details}

All experiments were conducted on a Tesla V100-PCIE-16GB GPU using PyTorch 2.6.0 with CUDA 12.4. The Transformers library (version 4.57.1) was used for loading DINOv3, and the PEFT library provided QLoRA and DoRA implementations. Quantization was performed using the bitsandbytes library.

Training hyperparameters for PEFT approaches were kept consistent across all configurations: 80 epochs, batch size 4 with gradient accumulation over 8 steps (effective batch size 32), learning rate $1 \times 10^{-4}$ with 3\% linear warmup followed by cosine annealing to 10\% of peak learning rate, and weight decay 0.01. Mixed precision training (automatic mixed precision with float16) was enabled for all PEFT experiments. Gradient checkpointing was enabled to reduce memory consumption at the cost of increased computation time.

Early stopping with patience of 5 epochs was implemented based on validation accuracy, and the best model checkpoint was saved for final evaluation. All experiments used a fixed random seed of 42 for reproducibility.

\subsection{Evaluation Protocol}

Model performance was evaluated using multiple metrics. Overall accuracy was used to measure the proportion of correctly classified samples across all classes. To account for class imbalance, the weighted F1-score was computed by weighting each class's F1-score by its support. In addition, per-class precision, recall, and F1-score were calculated to identify behaviors that were particularly easy or difficult to classify. Detailed classification reports and confusion matrices were generated using scikit-learn.

Evaluation was performed on two distinct sets. The validation set consisted of 540 samples (60 per class) randomly sampled from the curated dataset, used for hyperparameter selection and early stopping. The resulting evaluation protocol, in which the test set was approximately 98 times larger than the training set, provided a rigorous assessment of model generalization beyond the curated training distribution. To further analyze robustness across data sources, performance metrics were also computed separately for each source (MMCows and PlayBehavior) to assess cross-source generalization.

In addition to predictive performance, inference efficiency was characterized by two metrics: Latency, defined as the average processing time per image (milliseconds); and Throughput, defined as the rate of sample processing (images/second) under maximum GPU utilization (measured using batch size 32).

\section{Results}
\label{sec:results}

\subsection{Comparison of Training Approaches}

\begin{figure}[!htbp]
\centering
\includegraphics[width=\textwidth]{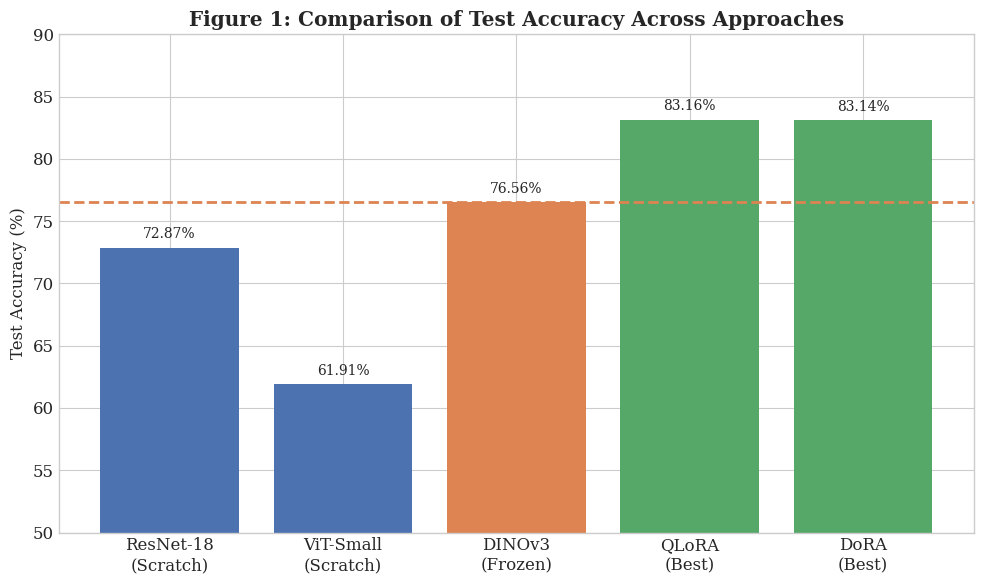}
\caption{Comparison of test accuracy across training approaches for dairy cow behavior classification. Four categories are shown: training from scratch (ResNet-18, ViT-Small), frozen feature extraction (DINOv3), and parameter-efficient fine-tuning methods (QLoRA, DoRA) with varying configurations. The horizontal dashed line indicates the frozen DINOv3 baseline performance (76.56\%).}
\label{fig:test_accuracy_comparison}
\end{figure}

Table~\ref{tab:main_results} presents the comprehensive performance comparison across all evaluated approaches. Parameter-efficient fine-tuning (PEFT) methods substantially outperformed both training from scratch and frozen feature extraction approaches. The best-performing configuration, QLoRA with all-linear target modules and rank=64, achieved 83.16\% test accuracy and 0.838 weighted F1-score, representing improvements of 10.29 and 6.60 percentage points over ResNet-18 (72.87\%) and frozen DINOv3 (76.56\%), respectively (Figure~\ref{fig:test_accuracy_comparison}).

\begin{table}[!htbp]
\centering
\caption{Performance comparison of all evaluated approaches for nine-class dairy cow behavior classification. Results are reported on the test set containing 211,800 samples. Training time was measured on a Tesla V100-PCIE-16GB GPU. The best results are highlighted in bold.}
\label{tab:main_results}
\small
\begin{tabular}{llccccc}
\toprule
\textbf{Method} & \textbf{Target} & \textbf{Rank} & \textbf{Trainable Params} & \textbf{Training Time} & \textbf{Test Acc.} & \textbf{Test F1} \\
\midrule
ResNet-18 (scratch) & --- & --- & 11.2M (100\%) & 16h 45m & 72.87\% & 0.7526 \\
ViT-Small (scratch) & --- & --- & 21.7M (100\%) & 18h 39m & 61.91\% & 0.6600 \\
DINOv3 (frozen) & --- & --- & 4.7M (0.07\%) & 17h 27m & 76.56\% & 0.7691 \\
\midrule
QLoRA & q\_proj & 8 & 2.6M (0.04\%) & 6h 32m & 77.17\% & 0.7646 \\
QLoRA & q\_proj & 16 & 5.2M (0.08\%) & 7h 16m & 78.38\% & 0.7753 \\
QLoRA & all-linear & 16 & 46.8M (0.70\%) & 4h 43m & 80.40\% & 0.8069 \\
\textbf{QLoRA} & \textbf{all-linear} & \textbf{64} & \textbf{183.0M (2.72\%)} & \textbf{5h 46m} & \textbf{83.16\%} & \textbf{0.8380} \\
\midrule
DoRA & q\_proj & 8 & 2.8M (0.04\%) & 11h 31m & 81.53\% & 0.8182 \\
DoRA & q\_proj & 16 & 5.4M (0.08\%) & 10h 27m & 81.03\% & 0.8153 \\
DoRA & all-linear & 16 & 48.4M (0.72\%) & 11h 51m & 81.23\% & 0.8139 \\
\textbf{DoRA} & \textbf{all-linear} & \textbf{64} & \textbf{184.5M (2.75\%)} & \textbf{10h 59m} & \textbf{83.14\%} & \textbf{0.8338} \\
\bottomrule
\end{tabular}
\end{table}

Training from scratch yielded the lowest performance, with ResNet-18 achieving 72.87\% test accuracy and ViT-Small achieving only 61.91\%. Despite ViT-Small having nearly twice the parameters of ResNet-18 (21.7M vs.\ 11.2M), the convolutional architecture demonstrated superior performance when trained on limited data conditions, highlighting the data efficiency advantage of CNN-based models in this setting.

Frozen feature extraction using DINOv3 improved upon training from scratch by 3.69 percentage points (76.56\% vs.\ 72.87\%), indicating the value of large-scale pre-trained representations even when the backbone remains fixed and no task-specific adaptation is applied.

PEFT methods achieved the highest performance while requiring substantially less training time. QLoRA (all-linear, $r=64$) completed training in 5 hours and 46 minutes, compared to 16 hours and 45 minutes for ResNet-18 and 17 hours and 27 minutes for frozen DINOv3. This represents a 65.6\% reduction in training time relative to training from scratch, while simultaneously improving test accuracy by 10.29 percentage points.

\subsection{Effect of PEFT Hyperparameters}

\subsubsection{QLoRA Configuration Analysis}

Table~\ref{tab:qlora_configs} presents the systematic evaluation of QLoRA hyperparameters, revealing a consistent pattern: increasing adapter capacity improved both validation and test performance.

\begin{table}[!htbp]
\centering
\caption{Effect of QLoRA hyperparameters on model performance. The validation-test gap is calculated as the difference between validation and test accuracy.}
\label{tab:qlora_configs}
\small
\begin{tabular}{clccccc}
\toprule
\textbf{Config} & \textbf{Target Modules} & \textbf{Rank} & \textbf{Alpha} & \textbf{Trainable Params} & \textbf{Test Acc.} & \textbf{Val--Test Gap} \\
\midrule
Q1 & q\_proj     & 16 & 32  & 5.2M (0.08\%)   & 78.38\% & 12.18 pp \\
Q2 & q\_proj     & 8  & 16  & 2.6M (0.04\%)   & 77.17\% & 13.39 pp \\
Q3 & all-linear  & 16 & 32  & 46.8M (0.70\%)  & 80.40\% & 9.97 pp \\
Q4 & all-linear  & 64 & 128 & 183.0M (2.72\%) & 83.16\% & 8.14 pp \\
\bottomrule
\end{tabular}
\end{table}

Reducing rank from 16 to 8 (Q1$\to$Q2) decreased test accuracy from 78.38\% to 77.17\% while maintaining similar validation performance (90.56\%), suggesting that the generalization gap was not caused by overfitting due to excessive adapter capacity. Expanding target modules from \textit{q\_proj} to all-linear layers (Q2$\to$Q3) improved test accuracy by 3.23 percentage points, and further increasing rank to 64 (Q3$\to$Q4) yielded an additional 2.76 percentage point improvement.

\begin{figure}[!htbp]
\centering
\includegraphics[width=\textwidth]{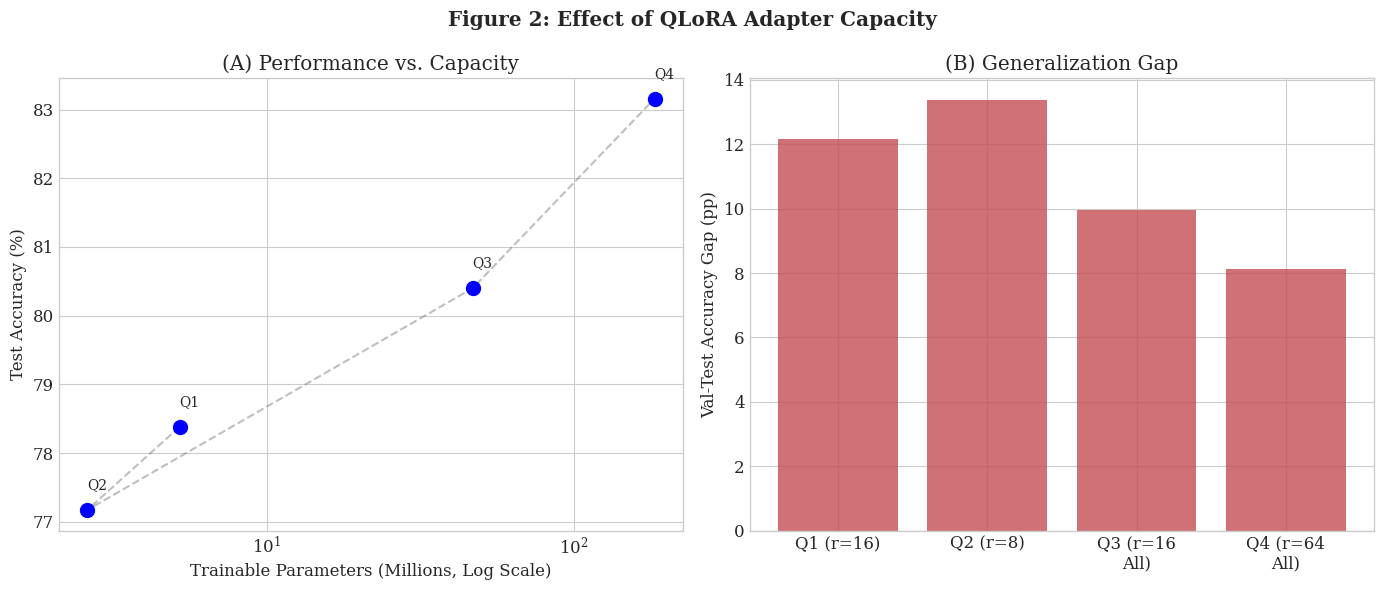}
\caption{Relationship between QLoRA adapter capacity and model performance. (A) Test accuracy as a function of trainable parameters. (B) Validation-test accuracy gap decreasing with increased adapter capacity. The trend demonstrates that increasing adapter capacity improved generalization rather than causing overfitting.}
\label{fig:qlora_capacity}
\end{figure}

The validation-test accuracy gap progressively decreased as adapter capacity increased: from 13.39 percentage points (Q2) to 8.14 percentage points (Q4). This inverse relationship between adapter capacity and generalization gap indicates that the observed discrepancy was primarily attributable to underfitting rather than overfitting (Figure~\ref{fig:qlora_capacity}).

\subsubsection{DoRA Configuration Analysis}

DoRA configurations exhibited similar patterns to QLoRA, with the all-linear, rank=64 configuration achieving the highest test accuracy (83.14\%) among DoRA variants (Table~\ref{tab:dora_configs}). Notably, DoRA demonstrated more stable performance across configurations compared to QLoRA, with test accuracy ranging from 81.03\% to 83.14\% (2.11 pp spread) compared to QLoRA's 77.17\% to 83.16\% (5.99 pp spread).

\begin{table}[!htbp]
\centering
\caption{Effect of DoRA hyperparameters on model performance.}
\label{tab:dora_configs}
\small
\begin{tabular}{clccccc}
\toprule
\textbf{Config} & \textbf{Target Modules} & \textbf{Rank} & \textbf{Alpha} & \textbf{Trainable Params} & \textbf{Test Acc.} & \textbf{Val--Test Gap} \\
\midrule
D1 & q\_proj     & 16 & 32  & 5.4M (0.08\%)   & 81.03\% & 10.64 pp \\
D2 & q\_proj     & 8  & 16  & 2.8M (0.04\%)   & 81.53\% & 8.66 pp \\
D3 & all-linear  & 16 & 32  & 48.4M (0.72\%)  & 81.23\% & 9.33 pp \\
D4 & all-linear  & 64 & 128 & 184.5M (2.75\%) & 83.14\% & 8.90 pp \\
\bottomrule
\end{tabular}
\end{table}

The \textit{q\_proj} configurations with DoRA (D1, D2) achieved higher test accuracy than their QLoRA counterparts (81.03--81.53\% vs.\ 77.17--78.38\%), suggesting that magnitude decomposition provided additional representational capacity that partially compensated for the limited target module coverage.

\subsubsection{QLoRA versus DoRA Comparison}

At equivalent configurations, QLoRA and DoRA achieved comparable final performance with the optimal settings (all-linear, $r=64$): 83.16\% vs.\ 83.14\% test accuracy, respectively. However, the two methods exhibited distinct characteristics in training efficiency and configuration sensitivity (Table~\ref{tab:qlora_vs_dora}).

\begin{table}[!htbp]
\centering
\caption{Direct comparison of QLoRA and DoRA at matched configurations. Training time ratio indicates DoRA time relative to QLoRA.}
\label{tab:qlora_vs_dora}
\small
\begin{tabular}{cccccccc}
\toprule
\textbf{Target} & \textbf{Rank} & \textbf{QLoRA Acc.} & \textbf{DoRA Acc.} & \textbf{$\Delta$} & \textbf{QLoRA Time} & \textbf{DoRA Time} & \textbf{Ratio} \\
\midrule
q\_proj     & 8  & 77.17\% & 81.53\% & +4.36 pp  & 6h 32m  & 11h 31m & 1.76$\times$ \\
q\_proj     & 16 & 78.38\% & 81.03\% & +2.65 pp  & 7h 16m  & 10h 27m & 1.44$\times$ \\
all-linear  & 16 & 80.40\% & 81.23\% & +0.83 pp  & 4h 43m  & 11h 51m & 2.51$\times$ \\
all-linear  & 64 & 83.16\% & 83.14\% & $-$0.02 pp & 5h 46m  & 10h 59m & 1.90$\times$ \\
\bottomrule
\end{tabular}
\end{table}

DoRA required approximately 1.4--2.5$\times$ longer training time across all configurations. The performance advantage of DoRA over QLoRA diminished as adapter capacity increased: DoRA outperformed QLoRA by 4.36 pp at \textit{q\_proj}, $r=8$, but the methods achieved essentially equivalent performance at all-linear, $r=64$.

\subsection{Per-Class Performance Analysis}

Table~\ref{tab:per_class} presents the per-class performance metrics for the best-performing model (QLoRA, all-linear, $r=64$) compared to alternative approaches. Classification performance varied substantially across behavior classes, with Lying achieving the highest F1-score (0.9914) and Walking the lowest (0.2609).

\begin{table}[!htbp]
\centering
\caption{Per-class performance comparison across training approaches. P = Precision, R = Recall, F1 = F1-score. The best F1-score for each class is highlighted in bold.}
\label{tab:per_class}
\small
\setlength{\tabcolsep}{3pt}
\begin{tabular}{lrcccccccccccc}
\toprule
 & & \multicolumn{3}{c}{\textbf{ResNet-18}} & \multicolumn{3}{c}{\textbf{ViT-Small}} & \multicolumn{3}{c}{\textbf{DINOv3 (frozen)}} & \multicolumn{3}{c}{\textbf{QLoRA (best)}} \\
\cmidrule(lr){3-5}\cmidrule(lr){6-8}\cmidrule(lr){9-11}\cmidrule(lr){12-14}
\textbf{Behavior} & \textbf{Support} & P & R & F1 & P & R & F1 & P & R & F1 & P & R & F1 \\
\midrule
Drinking        & 3,011   & .35 & .69 & .47 & .24 & .62 & .35 & .44 & .71 & .54 & .42 & \textbf{.79} & \textbf{.55} \\
Eat.\ head down & 30,952  & .82 & .35 & .49 & .66 & .31 & .42 & .87 & .29 & .44 & \textbf{.88} & \textbf{.55} & \textbf{.68} \\
Eat.\ head up   & 18,783  & .60 & .43 & .50 & .46 & .43 & .45 & .51 & .55 & .53 & \textbf{.75} & \textbf{.58} & \textbf{.66} \\
Lying           & 83,509  & .94 & .98 & .96 & .89 & .87 & .88 & .96 & .99 & .97 & \textbf{.99} & \textbf{.99} & \textbf{.99} \\
Standing        & 69,807  & .79 & .70 & .74 & .74 & .52 & .61 & .78 & .78 & .78 & \textbf{.79} & \textbf{.85} & \textbf{.82} \\
Walking         & 3,819   & .06 & .44 & .10 & .05 & .47 & .08 & .12 & .53 & .19 & \textbf{.17} & \textbf{.52} & \textbf{.26} \\
Front.\ pushing & 600     & .47 & .91 & .62 & .59 & .66 & .62 & .63 & .84 & .72 & .62 & .87 & \textbf{.73} \\
Galloping       & 575     & .43 & .57 & .49 & .57 & .61 & .59 & .71 & .55 & .62 & \textbf{.82} & .50 & \textbf{.62} \\
Leaping         & 744     & .60 & .40 & .48 & .56 & .55 & .56 & .58 & .61 & .59 & \textbf{.65} & \textbf{.67} & \textbf{.66} \\
\bottomrule
\end{tabular}
\end{table}

Behaviors could be categorized into three performance tiers. High-performance behaviors (F1 $> 0.80$) included Lying (0.99) and Standing (0.82), which together comprised 72.4\% of the test set. Medium-performance behaviors (F1 $= 0.55$--$0.75$) included Frontal pushing (0.73), Eating head down (0.68), Leap (0.66), Eating head up (0.66), Gallop (0.62), and Drinking (0.55). Low-performance behaviors (F1 $< 0.30$) included only Walking (0.26).

\begin{figure}[!htbp]
\centering
\includegraphics[width=\textwidth]{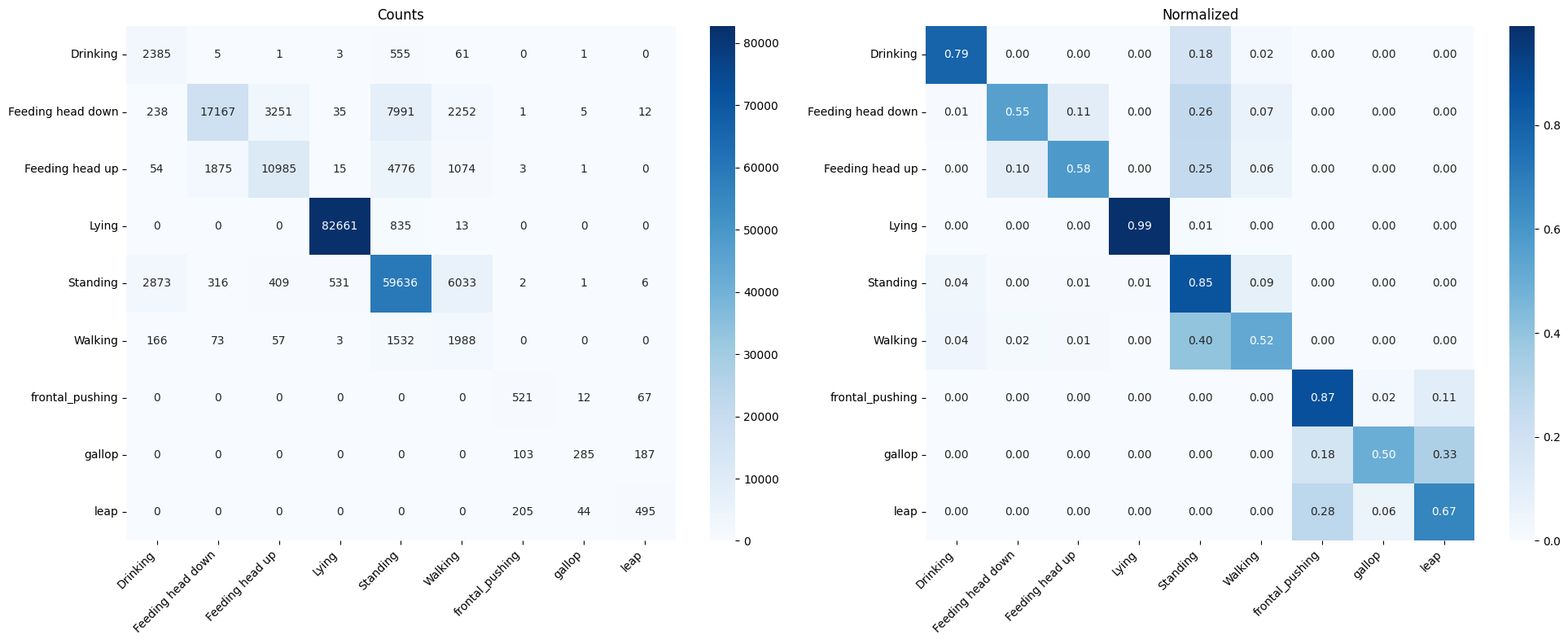}
\caption{Confusion matrix for the best-performing model (QLoRA, all-linear, $r=64$) on the test set ($n = 211{,}800$). Values are normalized by row to show the proportion of each true class assigned to each predicted class. The nine behavior classes are ordered by test set support from highest (Lying, $n = 83{,}509$) to lowest (Gallop, $n = 575$).}
\label{fig:confusion_matrix}
\end{figure}

The confusion matrix analysis (Figure~\ref{fig:confusion_matrix}) revealed systematic misclassification patterns. Walking was predominantly confused with Drinking (precision 0.17), reflecting the visual similarity of ambulatory behaviors. Eating head down and Eating head up exhibited bidirectional confusion (precision 0.88 and 0.75, respectively), attributable to the subtle postural differences between these eating states.

\subsection{Training Dynamics and Convergence}

\begin{figure}[!htbp]
\centering
\includegraphics[width=\textwidth]{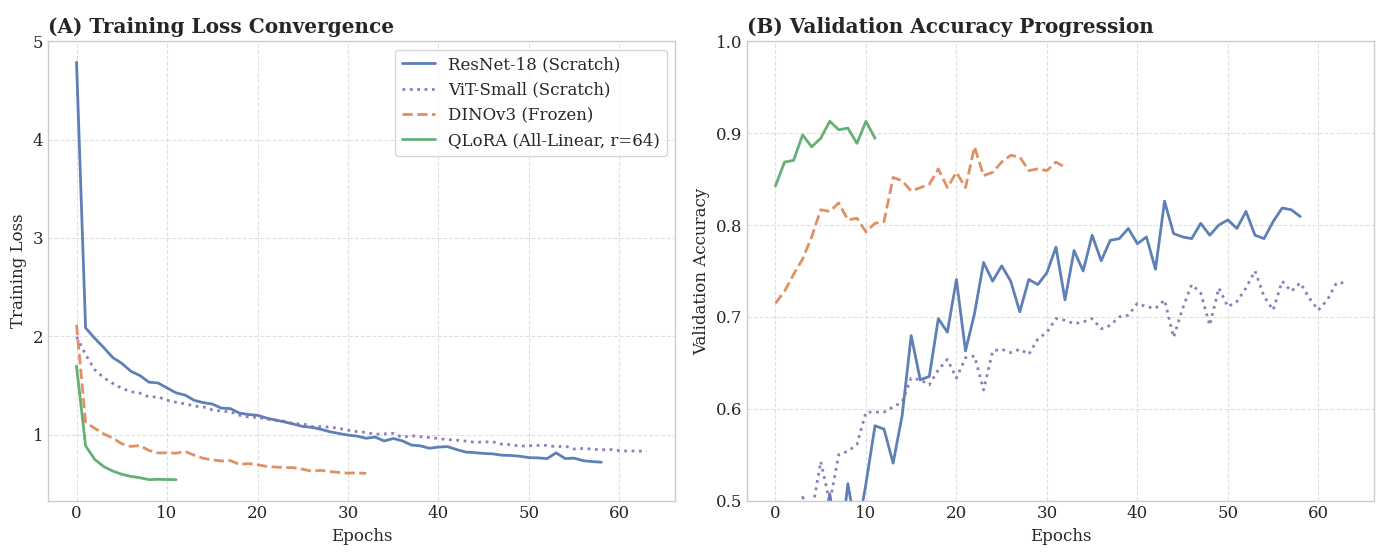}
\caption{Training dynamics for representative models from each approach. (A) Training loss curves showing convergence patterns across epochs. (B) Validation accuracy progression during training. Four approaches are shown: ResNet-18 trained from scratch (150 epochs), ViT-Small trained from scratch (150 epochs), frozen DINOv3 (80 epochs), and QLoRA all-linear $r=64$ (80 epochs).}
\label{fig:training_dynamics}
\end{figure}

Training curves for the best-performing configurations revealed distinct convergence patterns between training approaches. PEFT methods demonstrated rapid initial convergence followed by gradual refinement, while training from scratch exhibited slower but steady improvement throughout the training period (Figure~\ref{fig:training_dynamics}).

The frozen DINOv3 approach achieved high validation accuracy (88.52\%) rapidly but exhibited a 11.96 percentage point gap to test accuracy (76.56\%), indicating that the classification head alone could not adequately adapt the frozen features. In contrast, QLoRA (all-linear, $r=64$) achieved both the highest validation accuracy (91.30\%) and the smallest validation-test gap (8.14 pp), demonstrating superior generalization capability.

\subsection{Computational Efficiency}

Table~\ref{tab:efficiency} summarizes the computational requirements and inference efficiency across approaches.

\begin{table}[!htbp]
\centering
\caption{Computational efficiency comparison across training approaches. Inference time was measured on the complete test set ($n = 211{,}800$) using batch size 64 on a Tesla V100-PCIE-16GB GPU.}
\label{tab:efficiency}
\small
\begin{tabular}{llcccc}
\toprule
\textbf{Method} & \textbf{Configuration} & \textbf{Train Time} & \textbf{Inference Time} & \textbf{Throughput} & \textbf{Latency} \\
 & & & & (img/s) & (ms/img) \\
\midrule
ResNet-18 & scratch & 16h 45m & 16,124.9s & 13.1 & 76.13 \\
ViT-Small & scratch & 18h 39m & 14,975.5s & 14.1 & 70.70 \\
DINOv3 & frozen & 17h 27m & 31,848.2s & 6.6 & 150.36 \\
\midrule
QLoRA & q\_proj, $r\!=\!8$ & 6h 32m & 34,264.3s & 6.2 & 161.77 \\
QLoRA & q\_proj, $r\!=\!16$ & 7h 16m & 32,195.9s & 6.6 & 152.00 \\
QLoRA & all-linear, $r\!=\!16$ & 4h 43m & 31,962.4s & 6.6 & 150.90 \\
\textbf{QLoRA} & \textbf{all-linear, $r\!=\!64$} & \textbf{5h 46m} & \textbf{32,691.3s} & \textbf{6.5} & \textbf{154.34} \\
\midrule
DoRA & q\_proj, $r\!=\!8$ & 11h 31m & 31,903.7s & 6.6 & 150.63 \\
DoRA & q\_proj, $r\!=\!16$ & 10h 27m & 32,114.9s & 6.6 & 151.62 \\
DoRA & all-linear, $r\!=\!16$ & 11h 51m & 65,059.6s & 3.3 & 307.16 \\
\textbf{DoRA} & \textbf{all-linear, $r\!=\!64$} & \textbf{10h 59m} & \textbf{66,085.7s} & \textbf{3.2} & \textbf{312.01} \\
\bottomrule
\end{tabular}
\end{table}

\begin{figure}[!htbp]
\centering
\includegraphics[width=\textwidth]{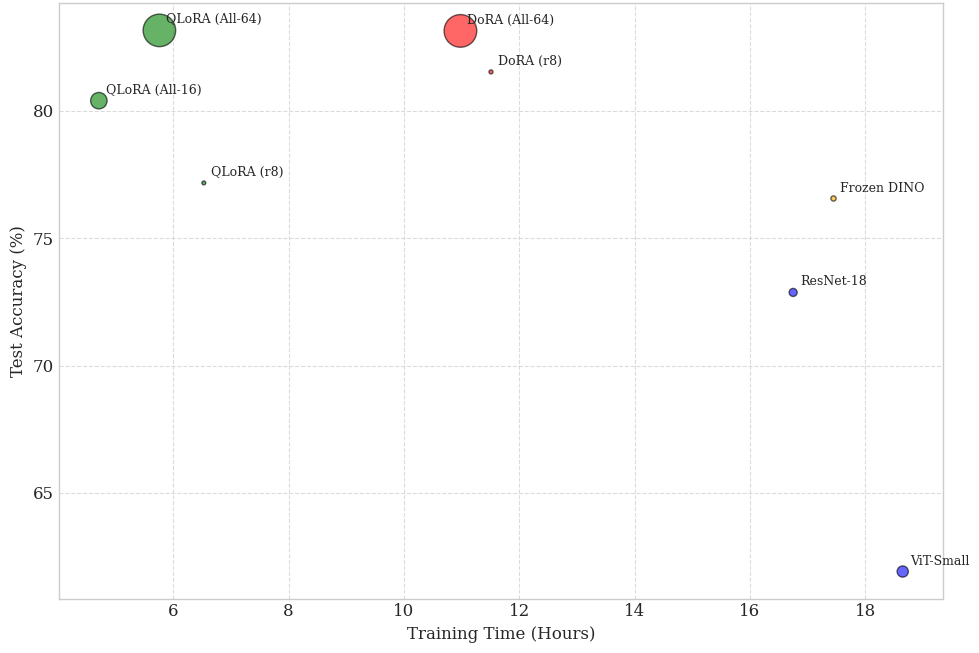}
\caption{Training efficiency versus model performance across all evaluated approaches. Each point represents a model configuration, with the x-axis showing training time in hours and the y-axis showing test accuracy. Point size indicates the number of trainable parameters. The Pareto frontier highlights QLoRA (all-linear, $r=64$) as achieving the best trade-off between training efficiency and model performance.}
\label{fig:efficiency}
\end{figure}

QLoRA achieved the fastest training times across all configurations. Inference throughput for DINOv3-based models ranged from 3.2 to 6.6 images per second, compared to 13.1--14.1 images per second for models trained from scratch. DoRA with all-linear configurations exhibited notably slower inference (3.2--3.3 img/s) compared to QLoRA (6.5--6.6 img/s), attributable to the additional magnitude computation overhead during forward passes (Figure~\ref{fig:efficiency}).

\section{Discussion}
\label{sec:discussion}

\subsection{Justification of Parameter-Efficient Fine-Tuning}

The results of this study provide compelling evidence for the advantages of using parameter-efficient fine-tuning (PEFT) over both training from scratch and frozen feature extraction for livestock behavior classification. The best QLoRA configuration achieved 83.16\% test accuracy, outperforming ResNet-18 trained from scratch (72.87\%) by 10.29 percentage points and frozen DINOv3 (76.56\%) by 6.60 percentage points. This performance advantage was achieved while simultaneously reducing training time by 65.6\% compared to ResNet-18 (5h 46m vs.\ 16h 45m) and 66.9\% compared to frozen DINOv3 (5h 46m vs.\ 17h 27m).

The efficiency gains of PEFT methods stem from their ability to leverage the rich representations encoded in pre-trained foundation models while adapting only a small subset of parameters to the target domain. In our experiments, the optimal QLoRA configuration updated only 2.72\% of the total model parameters (183.0M of 6.7B), yet this targeted adaptation was sufficient to achieve substantial performance improvements.

The frozen feature extraction approach achieved high validation accuracy (88.52\%) but exhibited a substantial 11.96 pp gap to test accuracy (76.56\%). It suggests that training only a classification head is insufficient to adapt DINOv3's pretrained representations to the domain-specific visual characteristics of livestock behavior imagery. PEFT methods address this limitation by enabling gradient flow through adapter modules distributed throughout the network, allowing the model to refine intermediate representations while preserving the general visual knowledge encoded in the frozen backbone.

These findings align with theoretical analyses suggesting that foundation models, despite their impressive zero-shot capabilities, often require domain-specific adaptation to achieve optimal performance on specialized tasks \cite{hu2022}. The agricultural domain presents unique visual challenges (e.g., variable lighting conditions, occlusions from barn infrastructure, and subtle postural differences between behavior classes) that may not be well-represented in general pre-training datasets. PEFT methods provide an effective mechanism for addressing this domain shift without the prohibitive computational costs of full fine-tuning.

From a practical deployment perspective, the reduced training time of PEFT methods has significant implications for agricultural AI applications. Livestock operations often require rapid model adaptation to new environments, camera configurations, or behavior classes of interest. The ability to fine-tune a state-of-the-art foundation model in under 6 hours on a single GPU makes iterative model refinement feasible even for operations with limited computational resources.

\subsection{Dataset Curation and Scalability}

A remarkable finding of this study is that 2,160 carefully curated training images were sufficient to achieve 83.16\% accuracy on a test set of 211,800 samples, corresponding to a 98:1 test-to-train ratio. This result demonstrates that dataset quality can substantially compensate for quantity when combined with appropriate transfer learning strategies.

The effectiveness of this small-data approach can be attributed to three main factors. First, rigorous manual verification ensured that training samples were correctly labeled and representative of their respective behavior classes, minimizing label noise. Second, a $3\times$ data augmentation strategy expanded the effective training set to 6,480 samples while introducing realistic visual variation. Third, the foundation model's pre-trained representations provided strong visual features that required only targeted refinement rather than learning from scratch.

During dataset preparation, however, we observed that some labels in the original datasets were incorrectly assigned. This prompted a targeted manual audit of 200 test images from the most frequently confused class pairs. The analysis showed that approximately 8\% were genuine model misclassifications, 32\% contained labels inconsistent with visible content, and 60\% depicted inherently ambiguous scenarios that could not be reliably classified from a single image. These findings suggest that the reported test accuracy may underestimate true model performance.

\begin{figure}[!htbp]
\centering
\includegraphics[width=0.65\textwidth]{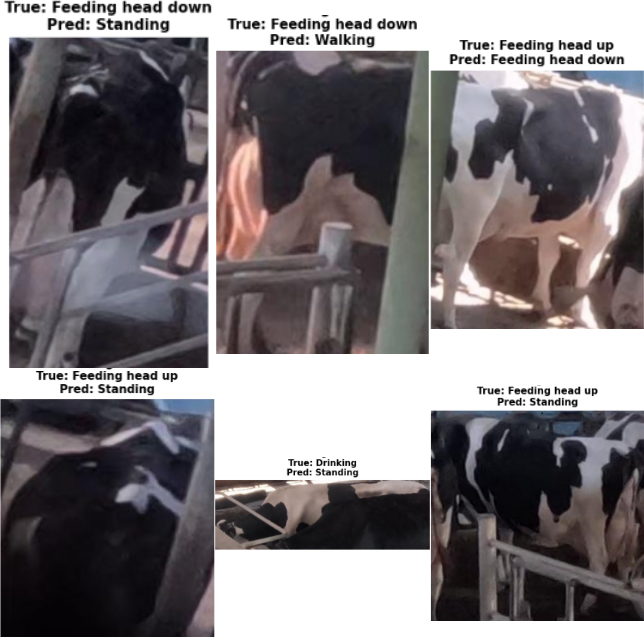}
\caption{Examples of test images with ground truth labels that are not visually determinable from the captured frame. Each sub-image displays the ground truth label (top) and the model prediction (bottom).}
\label{fig:label_issues1}
\end{figure}

\begin{figure}[!htbp]
\centering
\includegraphics[width=0.65\textwidth]{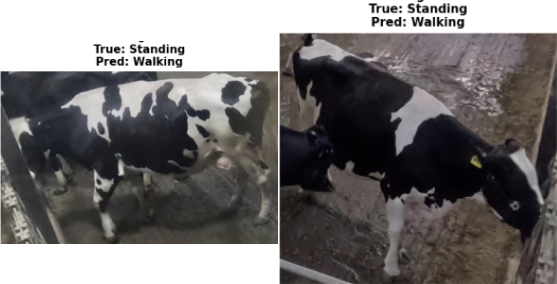}
\caption{Examples of images labeled as ``Standing'' that display limb configurations characteristic of locomotion.}
\label{fig:label_issues2}
\end{figure}

\begin{figure}[!htbp]
\centering
\includegraphics[width=0.8\textwidth]{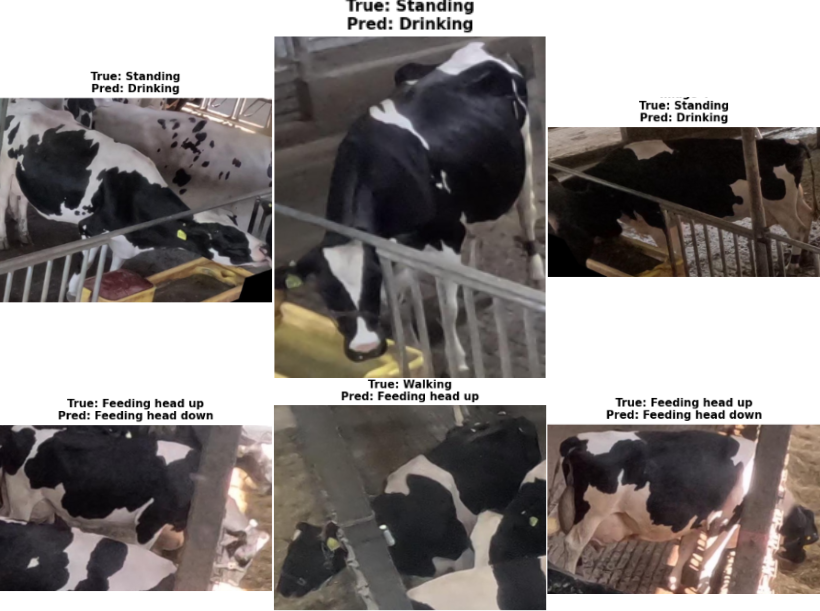}
\caption{Examples of apparent annotation errors in the test set. A cow positioned over a water trough is labeled as ``Standing'' rather than ``Drinking."}
\label{fig:label_issues3}
\end{figure}

\begin{figure}[!htbp]
\centering
\includegraphics[width=0.45\textwidth]{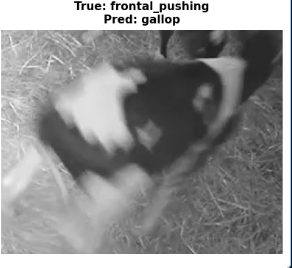}
\caption{Examples of compound behaviors in the PlayBehavior dataset. The model predicts the visually dominant behavior, whereas the ground truth label reflects a concurrent secondary action.}
\label{fig:label_issues4}
\end{figure}

The observed discrepancies arose from four principal sources (Figures~\ref{fig:label_issues1}--\ref{fig:label_issues4}): first, many labels referred to behaviors not visually determinable from the captured frame, such as hindquarters labeled as ``Eating head up,'' likely because annotations were derived from sensors capturing information inaccessible to cameras; second, some images showed limb configurations characteristic of locomotion yet were labeled as ``Standing''; third, clear annotation errors were observed, such as cows positioned over water troughs labeled as ``Standing'' rather than ``Drinking''; and fourth, certain images depicted compound behaviors, where a calf simultaneously galloping and performing frontal pushing was labeled with the secondary behavior.

Collectively, these findings highlight a broader challenge in precision livestock farming: the tension between sensor-derived behavioral annotations and the visual information available to computer vision systems. Cross-source analysis further revealed domain shift effects. Models achieved higher accuracy on MMCows (83.30\%) compared to PlayBehavior (67.80\%), indicating systematic visual differences between the datasets.

\subsection{Understanding the Underfitting Hypothesis}

One of the emerging insights from the experiments was the counterintuitive observation that increasing adapter capacity consistently improved generalization rather than causing overfitting. When the initial QLoRA configuration (\textit{q\_proj}, $r=16$) exhibited a large validation-test accuracy gap (12.18 pp), the natural assumption was overfitting. However, reducing rank from 16 to 8 decreased test accuracy from 78.38\% to 77.17\% while widening the gap to 13.39 pp. This unexpected result indicated underfitting: the adapter modules lacked sufficient capacity to learn the complex mapping required for robust behavior classification.

Subsequent experiments supported this hypothesis---expanding target modules from \textit{q\_proj} to all-linear layers improved test accuracy by 3.23 pp, and further increasing rank to 64 yielded an additional 2.76 pp improvement. The decreasing validation-test gap with increasing capacity (from 13.39 pp at \textit{q\_proj}, $r=8$ to 8.14 pp at all-linear, $r=64$) provided strong evidence; if overfitting were the issue, the gap would have widened instead.

Several factors explain why underfitting dominated this setting. First, adapting DINOv3's 6.7 billion parameters to the livestock domain may require substantial modifications that low-rank approximations cannot adequately capture. Second, the nine-class behavior classification task involves subtle visual distinctions that require learning fine-grained feature adjustments across multiple network layers. Third, the domain shift between DINOv3's pre-training distribution and agricultural imagery may necessitate more extensive adaptation than anticipated.

\subsection{QLoRA versus DoRA: When Does the Magnitude Matter?}

The comparison between QLoRA and DoRA revealed nuanced trade-offs that inform practical deployment decisions. At the optimal configuration (all-linear, $r=64$), both methods achieved virtually identical test accuracy (83.16\% vs.\ 83.14\%), suggesting that DoRA's magnitude decomposition did not translate to meaningful performance differences at high capacity. However, DoRA consistently outperformed QLoRA when adapter capacity was limited: at \textit{q\_proj}, $r=8$, DoRA achieved 81.53\% vs.\ QLoRA's 77.17\%, a difference of 4.36 pp.

This pattern reflects DoRA's architectural innovation. By decomposing pre-trained weights into magnitude and directional components \cite{liu2024dora}, DoRA provides additional representational flexibility that proves particularly beneficial under capacity constraints. When capacity is abundant, standard low-rank adaptation suffices, rendering the magnitude component redundant. However, DoRA required approximately 1.4--2.5$\times$ longer training time across all configurations. Based on these findings, we recommend: (1) QLoRA for high-capacity configurations (all-linear, $r \geq 64$) due to faster training with equivalent performance; (2) DoRA for low-capacity configurations (\textit{q\_proj} only or $r \leq 16$) where superior performance justifies additional training time.

\subsection{Comparison with Existing Literature}

Our results advance the state-of-the-art in livestock behavior recognition. The 83.16\% test accuracy represents competitive performance for multi-class behavior classification, particularly given the challenging evaluation protocol with a 98:1 test-to-train ratio.

Yang et al.\ \cite{yang2025pipeline} reported recall values exceeding 95\% for dairy cow behavior detection using DINOv2-based approaches, but evaluation was conducted on test sets drawn from the same distribution as training data. Similarly, Gao et al.\ \cite{gao2024} achieved 93.2\% detection accuracy with RFR-YOLO, though again on conventional train-test splits. Our large-scale generalization assessment provides a more stringent test of practical deployment capability.

The comparison with Espejo-Garc\'ia et al.\ \cite{espejo2025}, who applied DINOv2 with LoRA to crop disease and weed identification, is particularly relevant. Their study demonstrated accuracy improvements of 5--15 percentage points over training from scratch, consistent with our findings. However, their work did not systematically compare QLoRA and DoRA or investigate the effect of adapter capacity, making our study the first comprehensive evaluation of these PEFT variants for agricultural imagery.

The finding that increasing adapter capacity improved generalization contradicts prevailing assumptions in the PEFT literature, which typically emphasize the regularization benefits of low-rank constraints \cite{hu2022}. This discrepancy may reflect the unique characteristics of agricultural imagery, where the domain shift from pre-training distributions is particularly pronounced.

\subsection{Limitations}

Several limitations of this study should be acknowledged. First, the training images were randomly selected from the available dataset without targeted sampling for behavioral variation. Second, the dataset originated from specific farm environments, and generalization to fundamentally different settings was not evaluated. The performance gap between MMCows and PlayBehavior sources (15.50 pp) suggests that cross-domain generalization remains a challenge.

Third, the nine behavior classes represent a subset of behaviors relevant to precision livestock farming. Important behaviors such as estrus mounting, lameness indicators, and social interactions were not included. Fourth, inference latency presents a practical constraint for real-time deployment. The DINOv3-based models achieved throughput of 6.5 images per second compared to 13.1 for ResNet-18.

Fifth, all experiments were conducted on a single GPU (Tesla V100-PCIE-16GB), and scalability to multi-GPU training was not evaluated. Sixth, the study focused on image-based classification rather than video-based approaches that leverage temporal dynamics. Finally, the hyperparameter search, while systematic, was not exhaustive.

\subsection{Future Directions}

Several promising directions emerge from this study. First, improved dataset curation methodologies could enhance model performance on challenging behavior classes. Second, systematic hyperparameter optimization using automated methods could identify superior configurations beyond those evaluated. Third, multi-farm validation studies are essential for establishing generalizability. Fourth, real-time deployment optimization through model compression and knowledge distillation could address inference latency limitations. Fifth, integration with video-based approaches could improve performance on motion-dependent behaviors. Sixth, active learning frameworks could reduce annotation costs by intelligently selecting the most informative samples. Finally, integration with existing farm management systems presents opportunities for practical impact.

\section*{CRediT Authorship Contribution Statement}

\textbf{Haiyu Yang:} Conceptualization, Formal analysis, Methodology, Project administration, Resources, Software, Validation, Visualization, Writing -- original draft. \textbf{Sumit Sharma:} Data curation, Investigation, Writing -- original draft, Writing -- review \& editing. \textbf{Enhong Liu:} Writing -- review \& editing, Resources. \textbf{Miel Hostens:} Supervision, Writing -- review \& editing.

\section*{Declaration of Competing Interest}

The authors declare that they have no known competing financial interests or personal relationships that could have appeared to influence the work reported in this paper.

\section*{Data Availability}

The source code for all experiments is available at: \url{https://github.com/Bovi-analytics/PEFT-Fine-tuning-cows}. The pre-trained model weights for the best-performing configurations are available at: \url{https://huggingface.co/collections/Sonam5/peft4cows}. The MMCows dataset is publicly available at: \url{https://github.com/neis-lab/mmcows}. The PlayBehavior dataset is not publicly available yet. Due to the large size of the DINOv3 backbone (6.7B parameters), we provide only the trained LoRA adapter weights, which can be merged with the publicly available DINOv3 model from Hugging Face (\texttt{facebook/dinov3-vit7b16-pretrain-lvd1689m}).

\section*{Acknowledgments}

The authors gratefully acknowledge the creators of the MMCows dataset for making their data publicly available, which was instrumental to this study. We also acknowledge the use of the PlayBehavior dataset.


\appendix
\setcounter{table}{0}
\setcounter{figure}{0}
\renewcommand{\thetable}{S\arabic{table}}
\renewcommand{\thefigure}{S\arabic{figure}}

\section*{Supplementary Materials}

\subsection*{S1. Extended Hyperparameter Configurations}

\begin{table}[!htbp]
\centering
\caption{Complete hyperparameter configurations for all QLoRA experiments. All experiments used the DINOv3-ViT-Giant backbone (6.7B parameters) with 4-bit NormalFloat (NF4) quantization and double quantization enabled.}
\label{tab:s1_qlora}
\small
\begin{tabular}{lcccc}
\toprule
\textbf{Parameter} & \textbf{Q1} & \textbf{Q2} & \textbf{Q3} & \textbf{Q4} \\
 & (q\_proj, $r\!=\!16$) & (q\_proj, $r\!=\!8$) & (all-linear, $r\!=\!16$) & (all-linear, $r\!=\!64$) \\
\midrule
Target modules & q\_proj & q\_proj & all-linear & all-linear \\
Rank ($r$) & 16 & 8 & 16 & 64 \\
Alpha ($\alpha$) & 32 & 16 & 32 & 128 \\
LoRA dropout & 0.05 & 0.10 & 0.05 & 0.05 \\
Bias & none & none & none & none \\
Quantization type & NF4 & NF4 & NF4 & NF4 \\
Double quantization & True & True & True & True \\
Compute dtype & float16 & float16 & float16 & float16 \\
Epochs & 80 & 80 & 80 & 80 \\
Batch size & 4 & 4 & 4 & 4 \\
Gradient accumulation & 8 & 8 & 8 & 8 \\
Effective batch size & 32 & 32 & 32 & 32 \\
Learning rate & $1\times10^{-4}$ & $1\times10^{-4}$ & $1\times10^{-4}$ & $1\times10^{-4}$ \\
Weight decay & 0.01 & 0.01 & 0.01 & 0.01 \\
Warmup ratio & 0.03 & 0.03 & 0.03 & 0.03 \\
LR scheduler & Cosine & Cosine & Cosine & Cosine \\
Min LR ratio & 0.1 & 0.1 & 0.1 & 0.1 \\
Mixed precision & True & True & True & True \\
Gradient checkpointing & True & True & True & True \\
Trainable parameters & 5.2M & 2.6M & 46.8M & 183.0M \\
\% of total & 0.08\% & 0.04\% & 0.70\% & 2.72\% \\
\bottomrule
\end{tabular}
\end{table}

\begin{table}[!htbp]
\centering
\caption{Complete hyperparameter configurations for all DoRA experiments. DoRA configurations mirror QLoRA settings with the addition of \texttt{use\_dora=True} for magnitude decomposition.}
\label{tab:s2_dora}
\small
\begin{tabular}{lcccc}
\toprule
\textbf{Parameter} & \textbf{D1} & \textbf{D2} & \textbf{D3} & \textbf{D4} \\
 & (q\_proj, $r\!=\!16$) & (q\_proj, $r\!=\!8$) & (all-linear, $r\!=\!16$) & (all-linear, $r\!=\!64$) \\
\midrule
Target modules & q\_proj & q\_proj & all-linear & all-linear \\
Rank ($r$) & 16 & 8 & 16 & 64 \\
Alpha ($\alpha$) & 32 & 16 & 32 & 128 \\
use\_dora & True & True & True & True \\
LoRA dropout & 0.05 & 0.05 & 0.05 & 0.05 \\
Bias & none & none & none & none \\
Quantization type & NF4 & NF4 & NF4 & NF4 \\
Double quantization & True & True & True & True \\
Compute dtype & float16 & float16 & float16 & float16 \\
Epochs & 80 & 80 & 80 & 80 \\
Batch size & 4 & 4 & 4 & 4 \\
Gradient accumulation & 8 & 8 & 8 & 8 \\
Effective batch size & 32 & 32 & 32 & 32 \\
Learning rate & $1\times10^{-4}$ & $1\times10^{-4}$ & $1\times10^{-4}$ & $1\times10^{-4}$ \\
Weight decay & 0.01 & 0.01 & 0.01 & 0.01 \\
Warmup ratio & 0.03 & 0.03 & 0.03 & 0.03 \\
LR scheduler & Cosine & Cosine & Cosine & Cosine \\
Min LR ratio & 0.1 & 0.1 & 0.1 & 0.1 \\
Mixed precision & True & True & True & True \\
Gradient checkpointing & True & True & True & True \\
Trainable parameters & 5.4M & 2.8M & 48.4M & 184.5M \\
\% of total & 0.08\% & 0.04\% & 0.72\% & 2.75\% \\
\bottomrule
\end{tabular}
\end{table}

\begin{table}[!htbp]
\centering
\caption{Training configurations for baseline approaches (training from scratch and frozen feature extraction).}
\label{tab:s3_baselines}
\small
\begin{tabular}{lccc}
\toprule
\textbf{Parameter} & \textbf{ResNet-18} & \textbf{ViT-Small} & \textbf{DINOv3 (frozen)} \\
\midrule
Architecture & ResNet-18 & ViT-Small (patch16) & DINOv3-ViT-Giant \\
Pre-trained weights & None & None & facebook/dinov3-vit7b16 \\
Total parameters & 11.2M & 21.7M & 6.7B \\
Trainable parameters & 11.2M (100\%) & 21.7M (100\%) & 4.7M (0.07\%) \\
Feature dimension & 512 & 384 & 4096 \\
Classification head & Linear($512\!\to\!9$) & Linear($384\!\to\!9$) & MLP($4096\!\to\!1024\!\to\!512\!\to\!9$) \\
Head dropout & --- & --- & 0.30, 0.15 \\
Epochs & 150 & 150 & 80 \\
Batch size & 32 & 32 & 8 \\
Gradient accumulation & 1 & 1 & 4 \\
Effective batch size & 32 & 32 & 32 \\
Learning rate & $1\times10^{-3}$ & $5\times10^{-5}$ & $1\times10^{-3}$ \\
Weight decay & 0.01 & 0.01 & 0.01 \\
LR scheduler & Cosine & Cosine & Cosine \\
Label smoothing & 0.1 & 0.1 & 0.1 \\
Early stopping patience & 10 & 10 & 5 \\
Precision & float32 & float32 & float16 \\
\bottomrule
\end{tabular}
\end{table}

\subsection*{S2. Per-Class Confusion Matrices}

\begin{figure}[!htbp]
\centering
\includegraphics[width=0.7\textwidth]{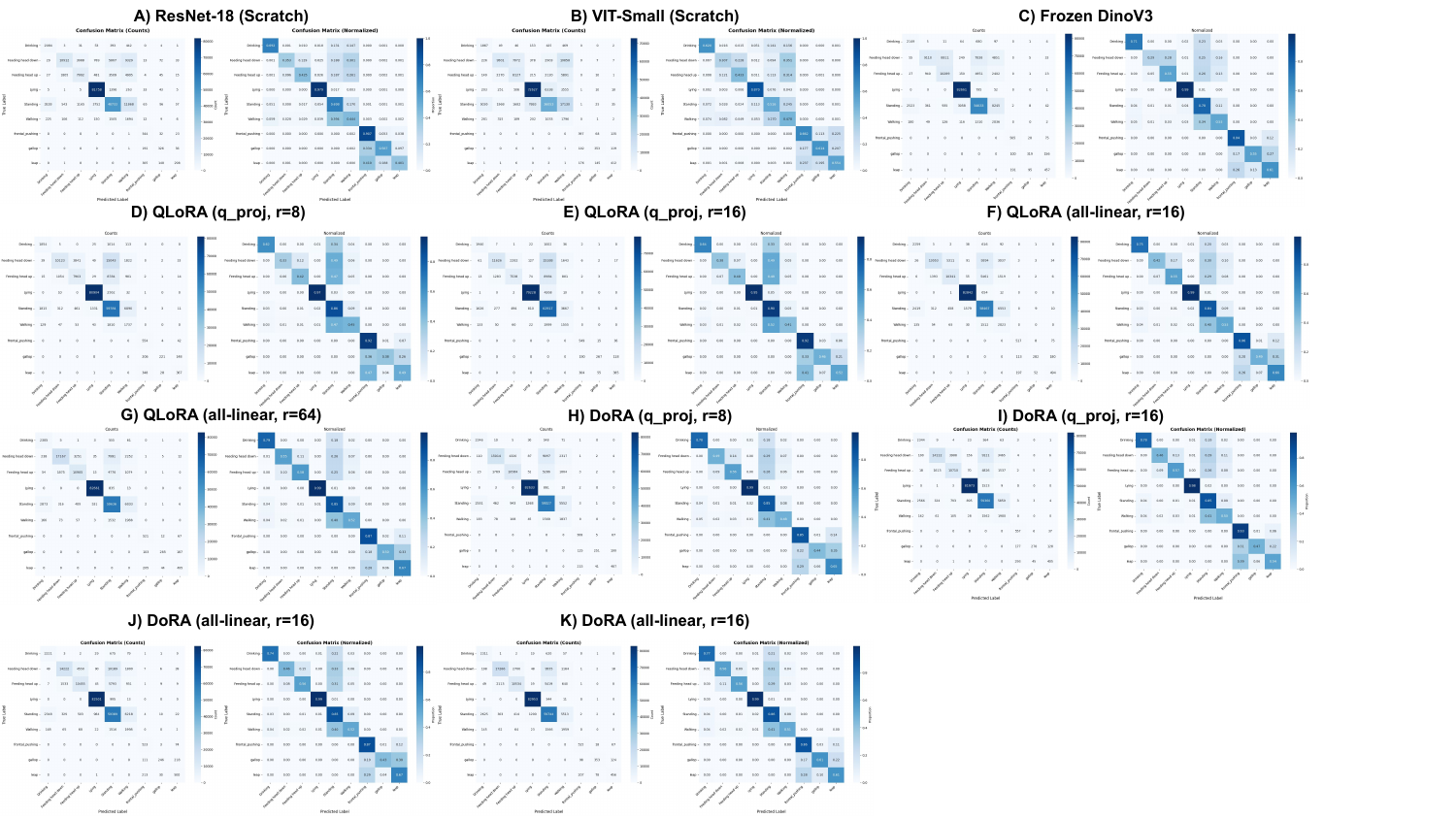}
\caption{Confusion matrices for all evaluated approaches on the test set ($n = 211{,}800$). Matrices are normalized by row to show recall for each class. (A) ResNet-18. (B) ViT-Small. (C) DINOv3 frozen. (D--G) QLoRA configurations Q1--Q4. (H--K) DoRA configurations D1--D4.}
\label{fig:s1_confusion}
\end{figure}

\begin{table}[!htbp]
\centering
\caption{Complete per-class performance metrics for all QLoRA configurations. P = Precision, R = Recall, F1 = F1-score.}
\label{tab:s4_qlora_perclass}
\small
\setlength{\tabcolsep}{2.5pt}
\begin{tabular}{lrcccccccccccc}
\toprule
 & & \multicolumn{3}{c}{\textbf{Q1}} & \multicolumn{3}{c}{\textbf{Q2}} & \multicolumn{3}{c}{\textbf{Q3}} & \multicolumn{3}{c}{\textbf{Q4}} \\
\cmidrule(lr){3-5}\cmidrule(lr){6-8}\cmidrule(lr){9-11}\cmidrule(lr){12-14}
\textbf{Behavior} & \textbf{Support} & P & R & F1 & P & R & F1 & P & R & F1 & P & R & F1 \\
\midrule
Drinking & 3,011 & .42 & .76 & .54 & .41 & .75 & .53 & .42 & .78 & .55 & .42 & .79 & .55 \\
Eat.\ h.d. & 30,952 & .87 & .38 & .53 & .86 & .35 & .50 & .88 & .48 & .62 & .88 & .55 & .68 \\
Eat.\ h.u. & 18,783 & .60 & .58 & .59 & .57 & .57 & .57 & .70 & .57 & .63 & .75 & .58 & .66 \\
Lying & 83,509 & .97 & .99 & .98 & .97 & .99 & .98 & .99 & .99 & .99 & .99 & .99 & .99 \\
Standing & 69,807 & .78 & .82 & .80 & .77 & .81 & .79 & .79 & .84 & .81 & .79 & .85 & .82 \\
Walking & 3,819 & .13 & .52 & .21 & .12 & .51 & .19 & .15 & .52 & .23 & .17 & .52 & .26 \\
Front.\ push. & 600 & .62 & .85 & .72 & .60 & .84 & .70 & .62 & .87 & .72 & .62 & .87 & .73 \\
Gallop & 575 & .75 & .50 & .60 & .72 & .48 & .58 & .79 & .49 & .61 & .82 & .50 & .62 \\
Leap & 744 & .60 & .62 & .61 & .58 & .60 & .59 & .63 & .65 & .64 & .65 & .67 & .66 \\
\midrule
\textbf{Weighted} & \textbf{211,800} & .82 & .78 & .78 & .81 & .77 & .76 & .85 & .80 & .81 & .86 & .83 & .84 \\
\bottomrule
\end{tabular}
\end{table}

\begin{table}[!htbp]
\centering
\caption{Complete per-class performance metrics for all DoRA configurations. P = Precision, R = Recall, F1 = F1-score.}
\label{tab:s5_dora_perclass}
\small
\setlength{\tabcolsep}{2.5pt}
\begin{tabular}{lrcccccccccccc}
\toprule
 & & \multicolumn{3}{c}{\textbf{D1}} & \multicolumn{3}{c}{\textbf{D2}} & \multicolumn{3}{c}{\textbf{D3}} & \multicolumn{3}{c}{\textbf{D4}} \\
\cmidrule(lr){3-5}\cmidrule(lr){6-8}\cmidrule(lr){9-11}\cmidrule(lr){12-14}
\textbf{Behavior} & \textbf{Support} & P & R & F1 & P & R & F1 & P & R & F1 & P & R & F1 \\
\midrule
Drinking & 3,011 & .43 & .77 & .55 & .43 & .77 & .55 & .43 & .77 & .55 & .43 & .77 & .55 \\
Eat.\ h.d. & 30,952 & .87 & .49 & .63 & .87 & .51 & .64 & .87 & .49 & .63 & .87 & .56 & .68 \\
Eat.\ h.u. & 18,783 & .72 & .57 & .64 & .74 & .56 & .64 & .73 & .56 & .63 & .77 & .56 & .65 \\
Lying & 83,509 & .98 & .99 & .99 & .98 & .99 & .99 & .98 & .99 & .99 & .98 & .99 & .99 \\
Standing & 69,807 & .78 & .84 & .81 & .79 & .84 & .81 & .78 & .84 & .81 & .77 & .86 & .81 \\
Walking & 3,819 & .17 & .51 & .26 & .18 & .52 & .27 & .18 & .51 & .27 & .21 & .51 & .30 \\
Front.\ push. & 600 & .62 & .86 & .72 & .62 & .86 & .72 & .62 & .86 & .72 & .63 & .86 & .72 \\
Gallop & 575 & .77 & .55 & .64 & .77 & .56 & .65 & .77 & .56 & .65 & .78 & .61 & .69 \\
Leap & 744 & .64 & .62 & .63 & .65 & .62 & .63 & .65 & .62 & .63 & .67 & .61 & .64 \\
\midrule
\textbf{Weighted} & \textbf{211,800} & .84 & .81 & .82 & .84 & .82 & .82 & .84 & .81 & .81 & .85 & .83 & .83 \\
\bottomrule
\end{tabular}
\end{table}

\subsection*{S3. Training Curves}

\begin{figure}[!htbp]
\centering
\includegraphics[width=\textwidth]{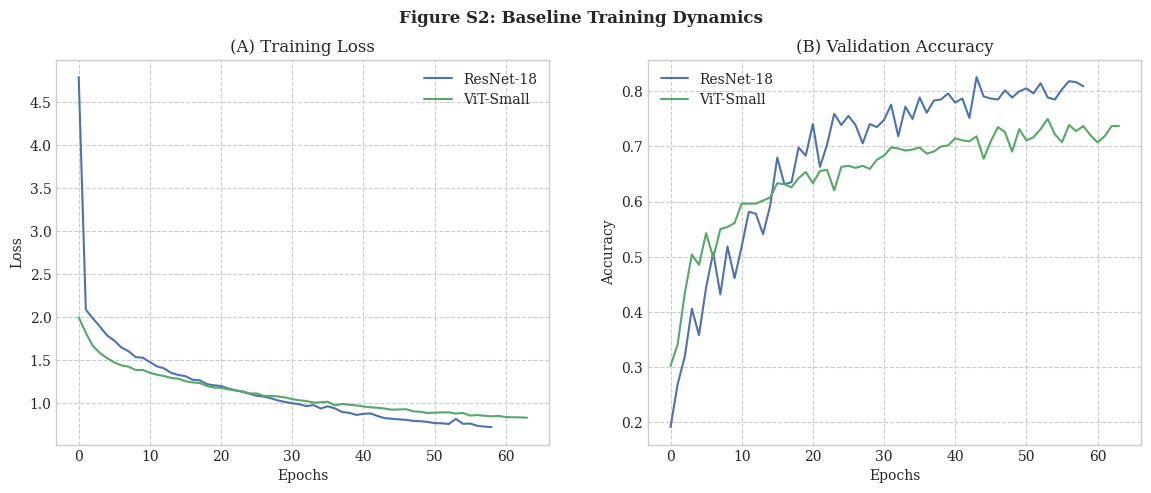}
\caption{Training dynamics for ResNet-18 and ViT-Small trained from scratch over 150 epochs. (A) Training loss. (B) Validation loss. (C) Training accuracy. (D) Validation accuracy. ResNet-18 achieved best validation accuracy of 75.93\% at epoch 142; ViT-Small achieved 62.78\% at epoch 127.}
\label{fig:s2_scratch}
\end{figure}

\begin{figure}[!htbp]
\centering
\includegraphics[width=\textwidth]{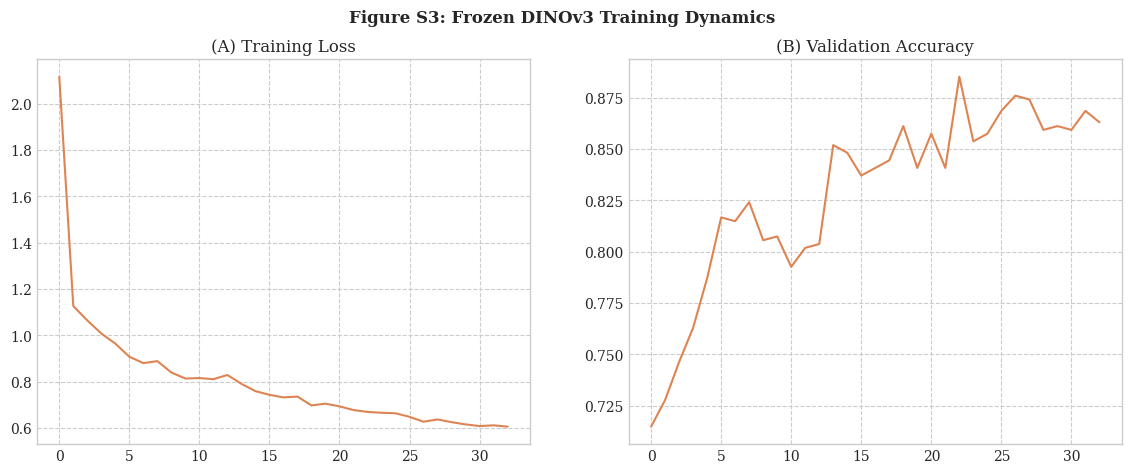}
\caption{Training dynamics for frozen DINOv3 feature extraction over 80 epochs. The model achieved best validation accuracy of 88.52\% at epoch 73.}
\label{fig:s3_frozen}
\end{figure}

\begin{figure}[!htbp]
\centering
\includegraphics[width=0.72\textwidth]{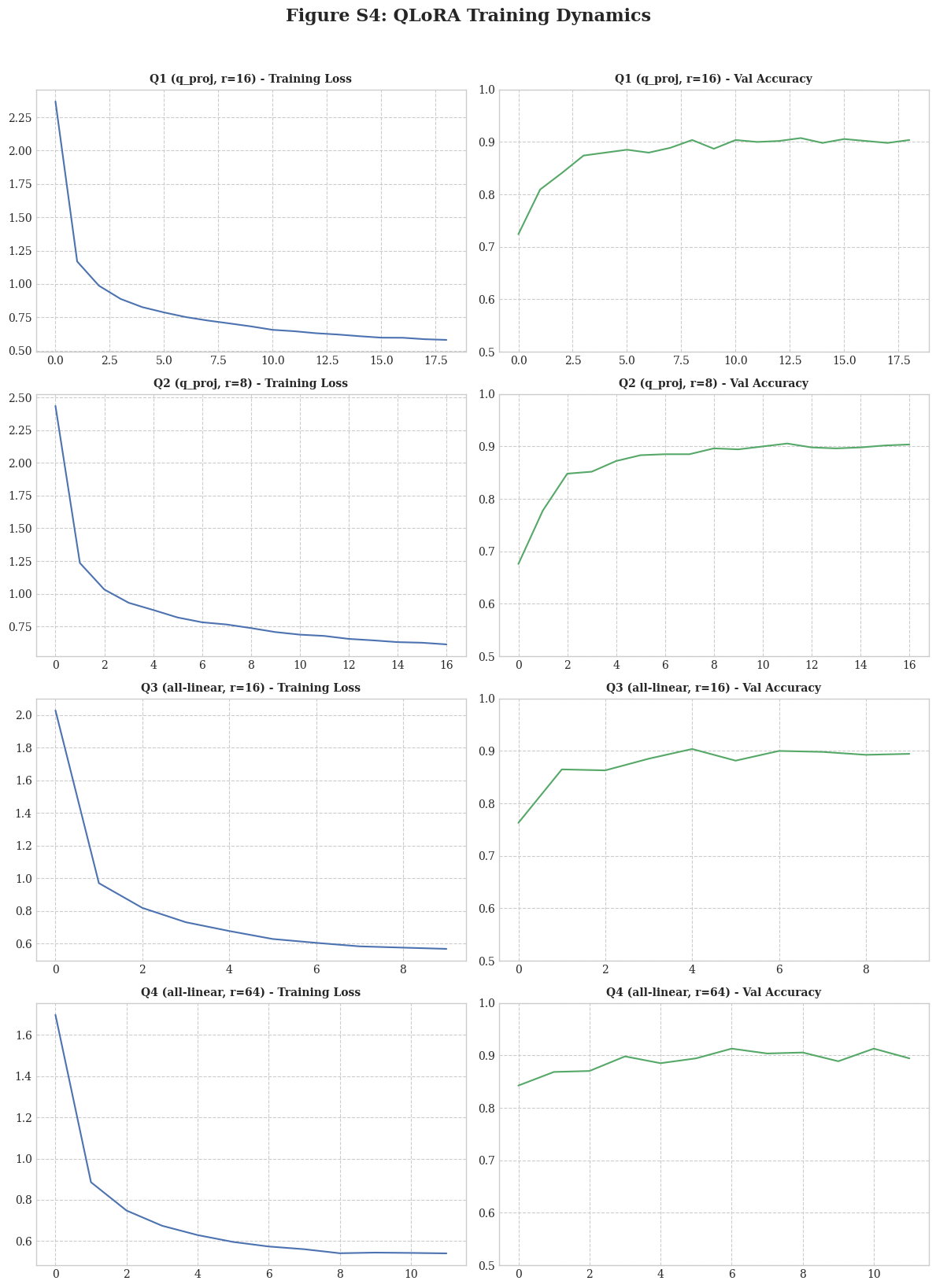}
\caption{Training dynamics for all QLoRA configurations over 80 epochs. Best validation accuracies: Q1=90.56\%, Q2=90.56\%, Q3=90.37\%, Q4=91.30\%.}
\label{fig:s4_qlora_curves}
\end{figure}

\begin{figure}[!htbp]
\centering
\includegraphics[width=0.72\textwidth]{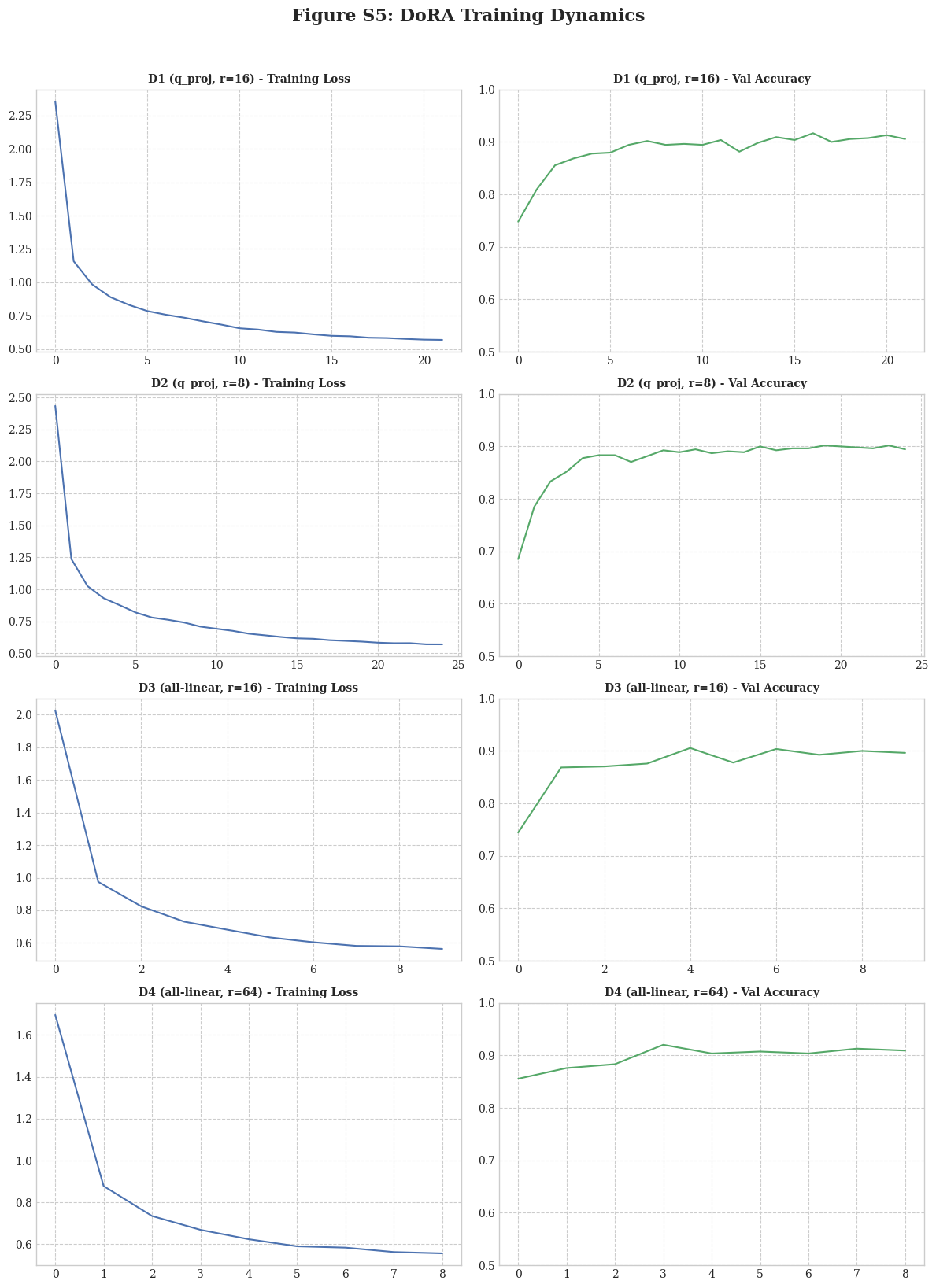}
\caption{Training dynamics for all DoRA configurations over 80 epochs. Best validation accuracies: D1=91.67\%, D2=90.19\%, D3=90.56\%, D4=92.04\%.}
\label{fig:s5_dora_curves}
\end{figure}

\begin{figure}[!htbp]
\centering
\includegraphics[width=\textwidth]{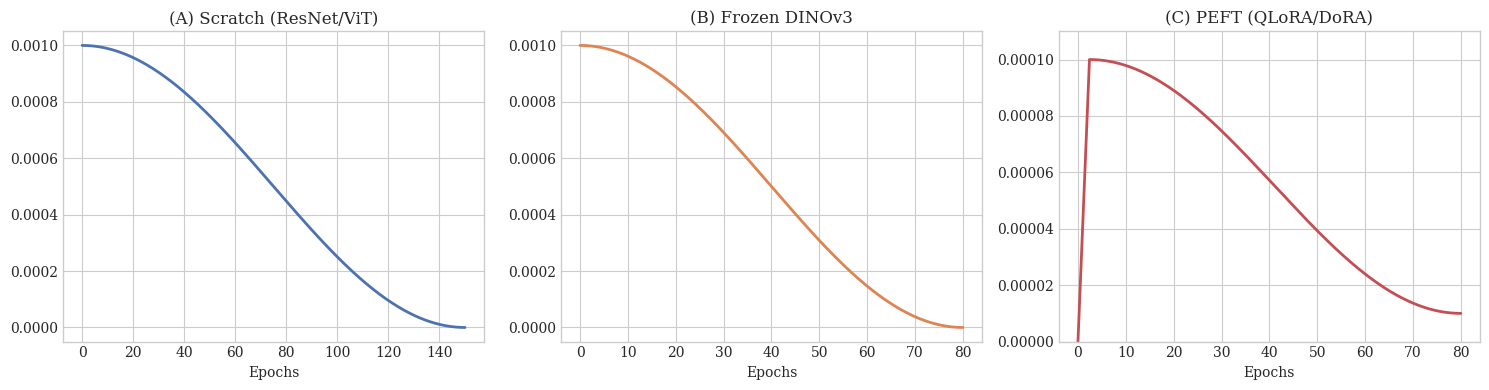}
\caption{Learning rate schedules for all training approaches. (A) Cosine annealing for ResNet-18 and ViT-Small (150 epochs). (B) Cosine annealing for frozen DINOv3 (80 epochs). (C) Cosine annealing with linear warmup (3\%) for all PEFT configurations (80 epochs).}
\label{fig:s6_lr}
\end{figure}

\subsection*{S4. Data Augmentation Pipeline}

\begin{table}[!htbp]
\centering
\caption{Complete data augmentation pipeline implemented using the Albumentations library. Augmentations were applied only during training with a $3\times$ multiplier.}
\label{tab:s6_augmentation}
\small
\begin{tabular}{llll}
\toprule
\textbf{Category} & \textbf{Transform} & \textbf{Parameters} & \textbf{Probability} \\
\midrule
\multirow{4}{*}{Geometric} & HorizontalFlip & --- & $p=0.5$ \\
 & Rotate & limit=$\pm15^{\circ}$ & $p=0.5$ \\
 & ShiftScaleRotate & shift=0.1, scale=(0.8, 1.2) & $p=0.5$ \\
 & Perspective & scale=(0.05, 0.10) & $p=0.3$ \\
\midrule
\multirow{3}{*}{Color (OneOf)} & RandomBrightnessContrast & $\pm$0.2 & \multirow{3}{*}{$p=0.8$ (group)} \\
 & HueSaturationValue & hue=10, sat=20, val=20 & \\
 & ColorJitter & bright=0.2, cont=0.2, sat=0.2, hue=0.1 & \\
\midrule
\multirow{3}{*}{Noise/Blur (OneOf)} & GaussNoise & var=(10, 50) & \multirow{3}{*}{$p=0.3$ (group)} \\
 & GaussianBlur & kernel=(3, 5) & \\
 & MotionBlur & kernel=5 & \\
\midrule
Occlusion & CoarseDropout & max\_holes=8, max\_size=10\% & $p=0.2$ \\
\midrule
\multirow{2}{*}{Normalization} & Resize & $224\times224$ & $p=1.0$ \\
 & Normalize & ImageNet mean/std & $p=1.0$ \\
\bottomrule
\end{tabular}
\end{table}

\begin{figure}[!htbp]
\centering
\includegraphics[width=\textwidth]{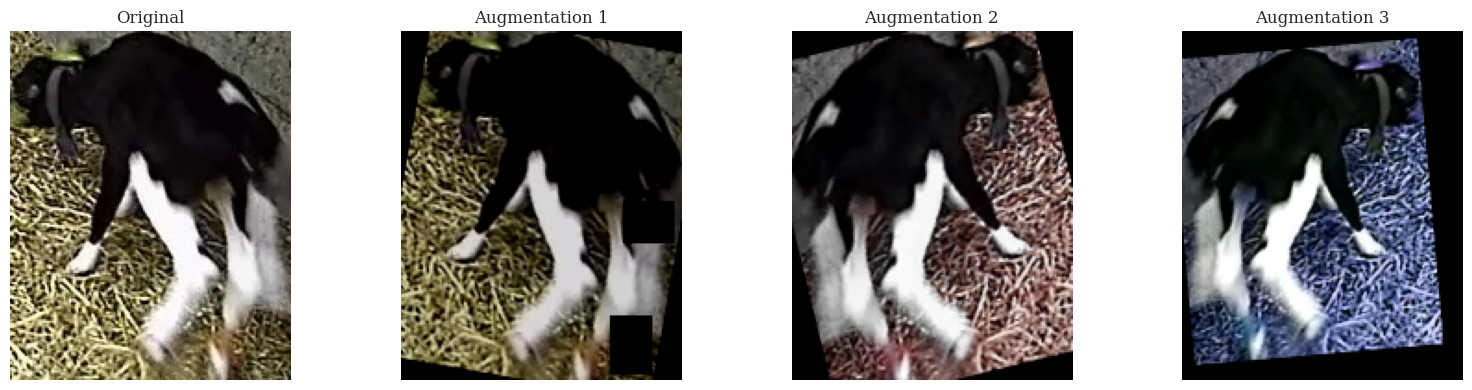}
\caption{Visualization of the data augmentation pipeline. Each row shows a single training image with its augmented versions. Nine examples are shown, one from each behavior class.}
\label{fig:s7_augmentation}
\end{figure}

\subsection*{S5. Dataset Details}

\begin{table}[!htbp]
\centering
\caption{Detailed breakdown of the dataset by behavior class and source.}
\label{tab:s7_dataset}
\small
\begin{tabular}{llrrrr}
\toprule
\textbf{Behavior} & \textbf{Source} & \textbf{Train} & \textbf{Val} & \textbf{Test} & \textbf{Total} \\
\midrule
Drinking        & MMCows       & 240 & 60 & 3,011   & 3,311   \\
Eating head down & MMCows      & 240 & 60 & 30,952  & 31,252  \\
Eating head up  & MMCows       & 240 & 60 & 18,783  & 19,083  \\
Lying           & MMCows       & 240 & 60 & 83,509  & 83,809  \\
Standing        & MMCows       & 240 & 60 & 69,807  & 70,107  \\
Walking         & MMCows       & 240 & 60 & 3,819   & 4,119   \\
Frontal pushing & PlayBehavior & 240 & 60 & 600     & 900     \\
Gallop          & PlayBehavior & 240 & 60 & 575     & 875     \\
Leap            & PlayBehavior & 240 & 60 & 744     & 1,044   \\
\midrule
\textbf{Total} & & \textbf{2,160} & \textbf{540} & \textbf{211,800} & \textbf{214,500} \\
\bottomrule
\end{tabular}
\end{table}

\begin{table}[!htbp]
\centering
\caption{Class imbalance analysis in the test set. The imbalance ratio is calculated relative to the smallest class (Gallop, $n=575$).}
\label{tab:s8_imbalance}
\small
\begin{tabular}{lrrc}
\toprule
\textbf{Behavior} & \textbf{Test Samples} & \textbf{Percentage} & \textbf{Imbalance Ratio} \\
\midrule
Lying           & 83,509 & 39.43\% & 145.2$\times$ \\
Standing        & 69,807 & 32.96\% & 121.4$\times$ \\
Eating head down & 30,952 & 14.61\% & 53.8$\times$ \\
Eating head up  & 18,783 & 8.87\%  & 32.7$\times$ \\
Walking         & 3,819  & 1.80\%  & 6.6$\times$ \\
Drinking        & 3,011  & 1.42\%  & 5.2$\times$ \\
Leap            & 744    & 0.35\%  & 1.3$\times$ \\
Frontal pushing & 600    & 0.28\%  & 1.0$\times$ \\
Gallop          & 575    & 0.27\%  & 1.0$\times$ \\
\bottomrule
\end{tabular}
\end{table}

\subsection*{S6. Classification Head Architecture}

\begin{table}[!htbp]
\centering
\caption{Architecture details of the classification head used for frozen DINOv3 and PEFT approaches.}
\label{tab:s9_head}
\small
\begin{tabular}{lcccc}
\toprule
\textbf{Layer} & \textbf{Input Dim} & \textbf{Output Dim} & \textbf{Activation} & \textbf{Dropout} \\
\midrule
Linear 1        & 4,096 & 1,024 & ReLU & 0.30 \\
Linear 2        & 1,024 & 512   & ReLU & 0.15 \\
Linear 3 (output) & 512 & 9     & ---  & ---  \\
\midrule
\multicolumn{4}{l}{\textbf{Total parameters}} & \textbf{4,719,113} \\
\bottomrule
\end{tabular}
\end{table}

The classification head was initialized with Kaiming uniform initialization for weights and zero initialization for biases. During training, the head parameters were always trainable regardless of whether the backbone was frozen or adapted via PEFT methods.

\subsection*{S7. Computational Environment}

\begin{table}[!htbp]
\centering
\caption{Software and hardware specifications for all experiments.}
\label{tab:s10_env}
\small
\begin{tabular}{ll}
\toprule
\textbf{Component} & \textbf{Specification} \\
\midrule
\multicolumn{2}{l}{\textit{Hardware}} \\
GPU & NVIDIA Tesla V100-PCIE-16GB \\
GPU Memory & 16 GB HBM2 \\
CPU & Intel Xeon (Azure Databricks) \\
System Memory & 56 GB \\
\midrule
\multicolumn{2}{l}{\textit{Software}} \\
Operating System & Ubuntu 22.04 LTS \\
Python & 3.10.12 \\
PyTorch & 2.6.0+cu124 \\
CUDA & 12.4 \\
Transformers & 4.57.1 \\
PEFT & 0.13.2 \\
bitsandbytes & 0.48.0 \\
Albumentations & 1.3.1 \\
scikit-learn & 1.3.2 \\
MLflow & 2.9.2 \\
\bottomrule
\end{tabular}
\end{table}

\subsection*{S8. Code and Data Availability}

The source code for all experiments, including data preprocessing, model training, and evaluation scripts, is available at: \url{https://github.com/Bovi-analytics/PEFT-Fine-tuning-cows}. The pre-trained model weights for the best-performing configurations (QLoRA all-linear $r=64$ and DoRA all-linear $r=64$) are available at: \url{https://huggingface.co/collections/Sonam5/peft4cows}. The MMCows dataset is publicly available at: \url{https://github.com/neis-lab/mmcows}. The PlayBehavior dataset is not publicly available yet.

\subsection*{S9. Reproducibility Checklist}

To ensure reproducibility, all experiments were conducted with the following controls:
\begin{itemize}
    \item \textbf{Random seed:} Fixed at 42 for all random operations (Python, NumPy, PyTorch, CUDA)
    \item \textbf{Deterministic operations:} CUDA deterministic mode enabled where possible
    \item \textbf{Data splits:} Train/validation/test splits saved and versioned
    \item \textbf{Model checkpoints:} Best model saved based on validation accuracy
    \item \textbf{Experiment tracking:} All runs logged to MLflow with full hyperparameter records
    \item \textbf{Version control:} All code versioned with Git
\end{itemize}

\noindent\textbf{Listing S1.} Random seed initialization code used across all experiments.
\begin{lstlisting}
import random
import numpy as np
import torch

def set_seed(seed: int = 42):
    """Set random seeds for reproducibility."""
    random.seed(seed)
    np.random.seed(seed)
    torch.manual_seed(seed)
    torch.cuda.manual_seed_all(seed)
    torch.backends.cudnn.deterministic = True
    torch.backends.cudnn.benchmark = False

set_seed(42)
\end{lstlisting}



\begin{thebibliography}{36}

\bibitem{berckmans2017}
D.~Berckmans, General introduction to precision livestock farming, Anim. Front. 7~(1) (2017) 6--11.

\bibitem{jiang2023}
M.~Jiang, Y.~Rao, J.~Zhang, Y.~Shen, Precision livestock farming research: A global scientometric review, Animals 13~(13) (2023) 2096.

\bibitem{rohan2024}
A.~Rohan, M.S.~Rafaq, M.J.~Hasan, F.~Asghar, A.K.~Bashir, T.~Dottorini, Application of deep learning for livestock behaviour recognition: A systematic literature review, Comput. Electron. Agric. 224 (2024) 109115.

\bibitem{zhang2023}
H.~Zhang, Y.~Sun, C.~Zhao, B.~Wang, B.~Li, B.~Wang, Review on typical behavior monitoring and physiological condition identification methods for ruminant livestock, Trans. Chin. Soc. Agric. Mach. 54 (2023) 1--21.

\bibitem{vannuffel2015}
A.~Van~Nuffel, I.~Zwertvaegher, L.~Pluym, S.~Van~Weyenberg, V.M.~Thorup, M.~Pastell, B.~Sonck, W.~Saeys, Lameness detection in dairy cows: Part~1, Animals 5~(3) (2015) 838--860.

\bibitem{robcis2023}
R.~Robcis, A.~Ferchiou, M.~Berrada, Y.~Ndiaye, N.~Herman, G.~Lhermie, D.~Raboisson, Cost of lameness in dairy herds: An integrated bioeconomic modeling approach, J. Dairy Sci. 106~(4) (2023) 2519--2534.

\bibitem{rollin2015}
E.~Rollin, K.C.~Dhuyvetter, M.W.~Overton, The cost of clinical mastitis in the first 30 days of lactation, Prev. Vet. Med. 122 (2015) 257--264.

\bibitem{liang2017}
D.~Liang, L.M.~Arnold, C.J.~Stowe, R.J.~Harmon, J.M.~Bewley, Estimating US dairy clinical disease costs with a stochastic simulation model, J. Dairy Sci. 100 (2017) 1472--1486.

\bibitem{rutten2013}
C.J.~Rutten, A.G.J.~Velthuis, W.~Steeneveld, H.~Hogeveen, Invited review: Sensors to support health management on dairy farms, J. Dairy Sci. 96 (2013) 1928--1952.

\bibitem{weigele2018}
H.C.~Weigele, L.~Gygax, A.~Steiner, B.~Wechsler, J.-B.~Burla, Moderate lameness leads to marked behavioral changes in dairy cows, J. Dairy Sci. 101 (2018) 2370--2382.

\bibitem{warner2020}
D.~Warner, E.~Vasseur, D.M.~Lefebvre, R.~Lacroix, A machine learning based decision aid for lameness in dairy herds using farm-based records, Comput. Electron. Agric. 169 (2020) 105193.

\bibitem{porto2015}
S.M.C.~Porto, C.~Arcidiacono, U.~Anguzza, G.~Cascone, The automatic detection of dairy cow feeding and standing behaviours in free-stall barns by a computer vision-based system, Biosyst. Eng. 133 (2015) 46--55.

\bibitem{wang2020}
S.H.~Wang, D.J.~He, D.~Liu, Automatic recognition method of dairy cow estrus behavior based on machine vision, Trans. Chin. Soc. Agric. Mach. 51~(4) (2020) 241--249.

\bibitem{qiao2022}
Y.~Qiao, Y.~Guo, K.~Yu, D.~He, C3D-ConvLSTM based cow behaviour classification using video data for precision livestock farming, Comput. Electron. Agric. 193 (2022) 106650.

\bibitem{fuentes2020}
A.~Fuentes, S.~Yoon, J.~Park, D.S.~Park, Deep learning-based hierarchical cattle behavior recognition with spatio-temporal information, Comput. Electron. Agric. 177 (2020) 105627.

\bibitem{yang2025pipeline}
H.~Yang, E.~Liu, J.~Sun, S.~Sharma, M.~van~Leerdam, S.~Franceschini, P.~Niu, M.~Hostens, A computer vision pipeline for individual-level behavior analysis: Benchmarking on the Edinburgh Pig Dataset, arXiv preprint arXiv:2509.12047, 2025.

\bibitem{bommasani2021}
R.~Bommasani, D.A.~Hudson, E.~Adeli, et~al., On the opportunities and risks of foundation models, arXiv preprint arXiv:2108.07258, 2021.

\bibitem{dosovitskiy2021}
A.~Dosovitskiy, L.~Beyer, A.~Kolesnikov, et~al., An image is worth 16x16 words: Transformers for image recognition at scale, in: Proc. ICLR, 2021.

\bibitem{caron2021}
M.~Caron, H.~Touvron, I.~Misra, H.~J\'egou, J.~Mairal, P.~Bojanowski, A.~Joulin, Emerging properties in self-supervised vision transformers, in: Proc. ICCV, 2021, pp.~9650--9660.

\bibitem{oquab2024}
M.~Oquab, T.~Darcet, T.~Moutakanni, et~al., DINOv2: Learning robust visual features without supervision, Trans. Mach. Learn. Res. (TMLR), 2024.

\bibitem{simeoni2025}
O.~Sim\'eoni, H.V.~Vo, M.~Seitzer, F.~Baldassarre, M.~Oquab, C.~Jose, A.~Joulin, P.~Bojanowski, DINOv3, arXiv preprint arXiv:2508.10104, 2025.

\bibitem{wortsman2022}
M.~Wortsman, G.~Ilharco, J.W.~Kim, et~al., Robust fine-tuning of zero-shot models, in: Proc. CVPR, 2022, pp.~7959--7971.

\bibitem{kumar2022}
A.~Kumar, A.~Raghunathan, R.~Jones, T.~Ma, P.~Liang, Fine-tuning can distort pretrained features and underperform out-of-distribution, in: Proc. ICLR, 2022.

\bibitem{andrew2021}
W.~Andrew, C.~Greatwood, T.~Burghardt, Visual localisation and individual identification of Holstein Friesian cattle via deep learning, in: Proc. ICCV Workshops, 2021, pp.~2850--2859.

\bibitem{bello2020}
R.W.~Bello, A.S.A.~Mohamed, A.Z.~Talib, Fine-tuning pre-trained models for automatic cashew disease classification, in: IEEE ICBDA, 2020, pp.~78--82.

\bibitem{houlsby2019}
N.~Houlsby, A.~Giurgiu, S.~Jastrzebski, et~al., Parameter-efficient transfer learning for NLP, in: Proc. ICML, 2019, pp.~2790--2799.

\bibitem{hu2022}
E.J.~Hu, Y.~Shen, P.~Wallis, Z.~Allen-Zhu, Y.~Li, S.~Wang, L.~Wang, W.~Chen, LoRA: Low-rank adaptation of large language models, in: Proc. ICLR, 2022.

\bibitem{dettmers2023}
T.~Dettmers, A.~Pagnoni, A.~Holtzman, L.~Zettlemoyer, QLoRA: Efficient finetuning of quantized LLMs, in: NeurIPS, vol.~36, 2023, pp.~10088--10115.

\bibitem{liu2024dora}
S.Y.~Liu, C.Y.~Wang, H.~Yin, P.~Molchanov, Y.C.F.~Wang, K.T.~Cheng, M.H.~Chen, DoRA: Weight-decomposed low-rank adaptation, in: Proc. ICML, 2024.

\bibitem{espejo2025}
B.~Espejo-Garcia, R.~G\"uldenring, L.~Nalpantidis, S.~Fountas, Foundation vision models in agriculture: DINOv2, LoRA and knowledge distillation for disease and weed identification, Comput. Electron. Agric. 239 (2025) 110900.

\bibitem{buslaev2020}
A.~Buslaev, V.I.~Iglovikov, E.~Khvedchenya, A.~Parinov, M.~Druzhinin, A.A.~Kalinin, Albumentations: Fast and flexible image augmentations, Information 11~(2) (2020) 125.

\bibitem{he2016}
K.~He, X.~Zhang, S.~Ren, J.~Sun, Deep residual learning for image recognition, in: Proc. CVPR, 2016, pp.~770--778.

\bibitem{wightman2019}
R.~Wightman, PyTorch Image Models, GitHub repository, 2019.

\bibitem{loshchilov2019}
I.~Loshchilov, F.~Hutter, Decoupled weight decay regularization, in: Proc. ICLR, 2019.

\bibitem{yang2026play}
H.~Yang, H.~Lesscher, E.~Liu, M.~Hostens, From manual observation to automated monitoring: Space allowance effects on play behaviour in group-housed dairy calves, 2026.

\bibitem{gao2024}
R.~Gao, H.~Chen, Y.~Zhang, S.~Wang, RFR-YOLO-based recognition method for dairy cow behavior in farming environments, Agriculture 15~(18) (2024) 1952.

\end{thebibliography}
\end{document}